\documentclass[11pt]{article}

\usepackage[final]{acl}

\usepackage{times}
\usepackage{latexsym}
\usepackage{titlesec}
\usepackage[T1]{fontenc}
\usepackage[utf8]{inputenc}

\usepackage{microtype}
\microtypesetup{patch=none}
\usepackage{inconsolata}

\usepackage{graphicx}

\usepackage{lipsum}

\usepackage{booktabs}

\usepackage{graphicx}
\usepackage{hyperref}
\usepackage{url}
\usepackage{makecell}
\definecolor{mylinkcolor}{RGB}{11, 20, 110}

\usepackage{amsmath}
\usepackage{amssymb}
\usepackage{mathtools}
\usepackage{amsthm}
\usepackage{adjustbox}
\usepackage{booktabs}

\usepackage{algorithm}  % Add this line
\usepackage{algorithmic}  % Add this line too for algorithm contents
\usepackage{cleveref}

\usepackage{wrapfig}  % Add this line for wrapped figures
\usepackage{caption}
\usepackage[table]{xcolor}

\usepackage{boxedminipage}
\usepackage{graphicx}  % For \resizebox command
\usepackage{colortbl}  % For \rowcolor command
\usepackage{multirow} 
\usepackage{xspace}
\usepackage[breakable]{tcolorbox}
\usepackage{titletoc} % table of contents
\usepackage{tabularx} % table of contents
\usepackage{subcaption}
\tcbuselibrary{listings}
\usepackage{listings}
\usepackage{multicol} % multicols
\usepackage{footnote} % enhanced footnote support
\usepackage[dvipsnames]{xcolor}
\usepackage{enumitem}
\usepackage{setspace}
\usepackage{soul}
\usepackage{wasysym} % provides \env
\usepackage{pifont} % provides many symbols

\usepackage{tcolorbox}
\tcbuselibrary{skins,breakable}
\usepackage{cuted}

\newcommand{\benchmark}{\textsc{ReviewBench}\xspace}
\newcommand{\method}{\textsc{ReviewGrounder}\xspace}

\makeatletter
\renewcommand{\paragraph}{\@startsection{paragraph}{4}{\z@}{0.0ex plus
0.3ex minus .2ex}{-1em}{\normalsize\textbf}}
\makeatother

\definecolor{myDodgerBlue}{RGB}{30, 144, 255}

\definecolor{darkgreen}{RGB}{0, 181, 18}
\definecolor{darkred}{RGB}{252, 90, 90}

\definecolor{DeltaBg}{HTML}{D4F2D7} % green
\definecolor{SearchBg}{HTML}{C2E6F5} % blue
\definecolor{AgenticBg}{HTML}{F5C2CC} % red
\definecolor{MathBg}{HTML}{E6D4F2} % purple
\definecolor{ScienceBg}{HTML}{FBE0BC} % yellow
\definecolor{warning}{RGB}{210, 40, 40}

\newcommand{\gain}[1]{$\uparrow$ #1}
\newcommand{\mygreen}[1]{\cellcolor{darkgreen!#1}}

\definecolor{toolcardbox}{RGB}{240, 248, 255} % Light blue background
\definecolor{toolcardborder}{RGB}{52, 52, 173} % Purple border

\newtcolorbox{promptbox}[2][]{
    colback=toolcardbox,
    colframe=toolcardborder,          
    coltitle=white,                
    arc=1pt,                       
    boxrule=1pt,
    fonttitle=\bfseries,
    title=#1,                      
    left=5pt,
    right=5pt,
    top=5pt,
    bottom=5pt,
    before skip=1em,
    after skip=1em,
    fontupper=\small,               
    breakable,     
    width=1.\linewidth, 
    #1                        
}

\newtcolorbox{reviewboxwide}[2][]{%
  enhanced,
  breakable,
  width=\textwidth,
  colback=black!3,
  colframe=brown!60!black,
  boxrule=0.8pt,
  arc=1.0mm,
  left=6pt,right=6pt,top=6pt,bottom=6pt,
  colbacktitle=brown!80!black,
  coltitle=white,
  fonttitle=\bfseries,
  title={#2},
  #1
}

\newtcolorbox{reviewquotetag}[1][]{%
  enhanced,
  breakable,
  colback=black!2,
  colframe=orange!60!black,
  boxrule=0.5pt,
  arc=1.0mm,
  left=6pt,right=6pt,top=4pt,bottom=4pt,
  before skip=4pt,
  after skip=6pt,
  colbacktitle=brown!80!black,
  coltitle=white,
  fonttitle=\bfseries,
  #1
}

\newcommand{\smalltt}[1]{{\ttfamily\fontsize{8}{10}\selectfont #1}}

\definecolor{mygreen}{RGB}{229, 245, 224}

\newlist{toolenum}{enumerate}{2}
\setlist[toolenum]{label=\arabic*., leftmargin=1.8em, itemsep=2pt, topsep=4pt}

\definecolor{codegray}{gray}{0.95} 

\newtcolorbox{simplecode}[1][]{%
    colback=toolcardbox,
    colframe=toolcardbox,
    arc=1pt,
    boxrule=0pt,
    top=-3pt, bottom=0pt,
    left=3pt, right=3pt,
    boxsep=0pt,
    left skip=0pt,
    right skip=0pt,
    fontupper=\ttfamily\fontsize{8.0}{10}\selectfont,
    breakable,
    #1
}

\newtcblisting{codebox}{
    colback=toolcardbox,
    colframe=white,
    arc=0pt,
    boxrule=0pt,
    listing only,
    listing options={
        language=Python,
        basicstyle=\ttfamily\footnotesize,
        keywordstyle=\color{blue!70!black}\bfseries,
        commentstyle=\itshape\color{green!40!black},
        stringstyle=\color{red!60!black},
        showstringspaces=false,
        columns=flexible,
        breaklines=true,
        inputencoding=utf8,
        extendedchars=true,
        texcl=true,
        escapeinside={(*@}{@*)},  % Define escape characters
    },
    left=0pt,
    right=0pt,
    top=-10pt,
    bottom=-10pt,
    breakable,
    fontupper=\small,
}

\newtheoremstyle{mythm}%
  {3pt}{3pt}{\itshape}{}{\bfseries}{.}{ }{}
\newtheoremstyle{mydef}%
  {3pt}{3pt}{}{}{\bfseries}{.}{ }{}

\theoremstyle{mydef}

\theoremstyle{mythm}

\title{\method: Improving Review Substantiveness with Rubric-Guided, Tool-Integrated Agents}

\author{
    \textbf{Zhuofeng Li}$^{1,*}$ \
    \textbf{Yi Lu}$^{2,*}$ \
    \textbf{Dongfu Jiang}$^{2}$ \
    \textbf{Haoxiang Zhang}$^{3}$ \
    \textbf{Yuyang Bai}$^{1}$ \\
    \textbf{Chuan Li}$^{4}$ \
    \textbf{Yu Wang}$^{5}$ \
    \textbf{Shuiwang Ji}$^{1}$ \
    \textbf{Jianwen Xie}$^{4,\dagger}$ \
    \textbf{Yu Zhang}$^{1,\dagger}$ \\
    $^{1}$Texas A\&M University \
    $^{2}$University of Waterloo \
    $^{3}$UC San Diego \
    $^{4}$Lambda \
    $^{5}$University of Oregon \\
}

\begin{document}
\maketitle
{\renewcommand{\thefootnote}{}\footnotetext{\textbf{*}: Equal Contribution. $\dagger$: Corresponding Author.}}

\begin{spacing}{1}
\begin{abstract}

{\color{warning} \textsc{Claim}: This work focuses on exploring how LLMs can assist human reviewers in the peer review process, rather than replacing them.}

The rapid rise in AI conference submissions has driven increasing exploration of large language models (LLMs) for peer review support.
However, LLM-based reviewers often generate superficial, formulaic comments lacking substantive, evidence-grounded feedback.
We attribute this to the underutilization of two key components of human reviewing: explicit rubrics and contextual grounding in existing work.
To address this, we introduce \benchmark, a benchmark evaluating review text according to paper-specific rubrics derived from official guidelines, the paper's content, and human-written reviews.
We further propose \method, a rubric-guided, tool-integrated multi-agent framework that decomposes reviewing into drafting and grounding stages, enriching shallow drafts via targeted evidence consolidation.
Experiments on \benchmark show that \method, using a Phi-4-14B-based drafter and a GPT-OSS-120B-based grounding stage, consistently outperforms baselines with substantially stronger/larger backbones (e.g., GPT-4.1 and DeepSeek-R1-670B) in both alignment with human judgments and rubric-based review quality across $8$ dimensions. The code is available \href{https://github.com/EigenTom/ReviewGrounder}{here}.
\end{abstract}

\section{Introduction}

Peer review is the primary mechanism through which the research community filters and improves new scientific work before publication.
The rapid growth of submissions at major AI conferences (with counts at leading venues surpassing 10,000) has placed sustained pressure on peer review workflows originally designed for far smaller scales~\cite{kim2025position}.
Meanwhile, recent advances in LLMs have spurred growing interest in using them to assist or complement the peer review workflow~\cite{zhang2024comprehensive}, for example, by drafting reviews~\cite{tan2024peer,zhu2025deepreview,zeng2025reviewrl}, summarizing reviewer opinions~\cite{du2024llms,hossain2025llms}, and providing feedback on review quality~\cite{feedback}.

Despite these advances, prior work has highlighted notable shortcomings in existing LLM-based peer review frameworks:
they produce routine, template-like critiques (e.g., ``\textit{add experiments on more data
sets}''; \citealp{liang2024can});
accept authors' claimed novelty or limitations without thorough verification~\cite{du2024llms,ye2024we};
and lack technical details, actionable suggestions, as well as justification grounded in the paper~\cite{zhou2024llm,du2024llms}.
Together, these limitations may lead to reviews that are \textbf{superficial and formulaic}, \textbf{lack substantive content and critical insights}, and prioritize syntax-level cues over the ability to deeply evaluate a paper's contributions.

Fundamentally, these shortcomings can be traced to the underutilization of two crucial sources of external information:
(1) \textbf{Reviewer Guidelines and Rubrics.} Top-tier NLP and machine learning venues~\cite{arr,icml_reviewer_instructions_2026,neurips_reviewer_guidelines_2025,iclr_reviewer_guide_2026} provide well-established peer-review guidelines that specify what to attend to in different review sections and which criteria to consider across evaluation dimensions. 
Compared with supervised fine-tuning solely on existing human-written reviews~\cite{zhu2025deepreview}, providing LLMs with clear, rubric-grounded instructions offers a more principled way to internalize how to produce substantive, content-rich reviews, especially since human reviews can be noisy and reviewers do not always follow official guidelines.
(2) \textbf{Context from Existing Work.} Reviewing should not be treated as a task that takes the submission alone as input.
In particular, assessing novelty inherently requires situating a paper relative to existing work.
When this context is absent, LLM-based reviewers have been observed to systematically underemphasize novelty when identifying weaknesses~\cite{shin2025mind}.
Addressing this limitation cannot be achieved by merely attaching retrieval-augmented generation~\cite{lewis2020retrieval}.
Instead, it requires a rubric-guided, tool-integrated, agentic framework with clear role separation (e.g., literature search, targeted section-level understanding, and rubric-guided synthesis) to support grounded evaluation.

\smallskip
\noindent \textbf{Contributions.}
In this paper, we aim to overcome the above shortcomings by explicitly targeting review substantiveness.
We first introduce \textbf{\benchmark}, an evaluation benchmark that leverages reviewer rubrics in an explicit and systematic manner.
\benchmark combines venue-provided generic guidelines with each paper's content and human-written reviews to instantiate paper-specific rubrics, and evaluates whether the generated review satisfies these requirements.
While agreement with human scores and decisions remains an important measure (and is therefore included), \benchmark shifts the focus toward what ultimately benefits authors and the community: actionable, rubric-grounded, and evidence-based feedback rather than the outcome alone.

Moreover, we propose \textbf{\method}, a rubric-guided, tool-integrated, multi-agent framework for producing grounded, content-rich reviews.
A single-pass review generator trained only on human-written reviews often produces shallow, mechanically structured drafts.
\method addresses this by decomposing reviewing into collaborating agents: the drafter produces an initial draft, and subsequent grounding agents refine it using tools for literature search, section-level analysis, evidence consolidation, and information aggregation. This process substantiates critiques, contextualizes novelty, and generates actionable suggestions.
Importantly, \method operates without paper-specific rubrics at generation time, ensuring improvements reflect deeper paper understanding rather than evaluation leakage.

We conduct a comprehensive evaluation of \method on \benchmark, measuring both review score and decision prediction alignment with human reviewers, as well as performance on $8$ rubric-specified dimensions (e.g., \textsc{Evidence-Based Critique}, \textsc{Constructive Tone}).
Across all tasks and metrics, \method with a Phi-4-14B-based drafter~\cite{abdin2024phi} and a GPT-OSS-120B-based grounding stage~\cite{agarwal2025gpt} consistently outperforms competitive baselines, including AI Scientist~\cite{lu2024ai}, AgentReview~\cite{jin2024agentreview}, CycleReviewer~\cite{weng2024cycleresearcher}, and DeepReviewer~\cite{zhu2025deepreview} with the same or even stronger/larger backbones, such as GPT-4o~\cite{hurst2024gpt}, GPT-4.1~\cite{openai_gpt41_2025}, and DeepSeek-R1~\cite{guo2025deepseek}.

The contributions of our work are as follows:
\begin{itemize}[leftmargin=*]
\item We identify review substantiveness as a key limitation of existing LLM-based reviewers, and introduce \benchmark, a rubric-driven benchmark that evaluates whether generated reviews provide accurate, evidence-grounded feedback beyond score or decision prediction.
\item To improve the substantiveness of reviews, we propose \method, a rubric-guided, tool-integrated multi-agent framework that decomposes the paper reviewing task into drafting and grounding stages, transforming shallow drafts into coherent and actionable reviews through explicit consolidation.
\item We conduct comprehensive experiments on \benchmark, which demonstrate that \method consistently produces more complete and more constructive reviews than competitive baselines, while achieving stronger alignment with human judgments.
\end{itemize}

\section{Related Work}
% Agentic System
% LLM-based Paper Review
% LLM-based Review Evaluation

% promising -> training-free -> training

% \noindent \textbf{Agentic System.} Agentic systems~\citep{wu2024autogen, lu2025octotools, jiang2025verltool,hu2024hireview,lu2025octotools} offer a promising way to solve complex science challenges. They consist of multiple modules, often including distinct LLMs assigned prescribed roles. By decomposing problems into sub-goals and iterating over multiple turns, these systems can tackle tasks that demand diverse tools, long horizons, or multi-stage reasoning. In this line of work, OWL~\citep{hu2025owl} introduces an extensible agentic framework and standardized tools for tackling complex reasoning tasks. Recently, AgentFlow~\citep{li2025flow} introduced a trainable, in-the-flow agentic framework that coordinates agent modules through an evolving memory and directly optimizes its planner inside the multi-turn loop. However, these advances largely target general problem solving, leaving the domain of academic peer review underexplored. Our framework addresses this gap by adapting agentic systems specifically for peer review.

\smallskip
\noindent \textbf{LLMs for Paper Review.} Recent studies~\citep{d2024marg,tan2024peer,du2024llms,hossain2025llms} have explored the use of LLMs to automate and enhance academic peer review.
For example, Reviewer2~\citep{gao2024reviewer2} proposes a two-stage framework that first generates aspect-specific prompts and then synthesizes reviews, improving both coverage and specificity;
AgentReview~\citep{jin2024agentreview} employs a multi-agent framework to simulate the peer review process.
More recently, DeepReview~\citep{zhu2025deepreview} is trained via supervised fine-tuning (SFT) on long chain-of-thought (CoT) data to enhance reasoning for review generation; ReviewRL~\citep{zeng2025reviewrl} introduces a reinforcement learning (RL) framework for producing scientific paper reviews. 
In practice, Review Feedback Agent~\citep{feedback} leverages multiple LLMs to improve review clarity and actionability at ICLR 2025.
Despite this progress, LLM-generated reviews often remain superficial and formulaic, reflecting the underutilization of explicit rubrics and contextual grounding in existing work and consequently struggling to provide substantive, evidence-grounded feedback.

\smallskip
\noindent \textbf{Automated Peer-Review Evaluation.} Despite growing interest in LLM-generated paper reviews, systematic evaluation frameworks remain scarce. Existing approaches can be broadly categorized into two types: (1) metric-based evaluation, where prior work~\citep{tan2024peer} relies on surface-level text similarity metrics such as ROUGE~\citep{lin2004rouge} and BLEU~\citep{papineni2002bleu}, or regression metrics like Mean Absolute Error (MAE); and (2) LLM-as-a-Judge evaluation~\citep{zhu2025deepreview}, which employs LLMs to directly assess generated reviews. However, these approaches fail to adequately assess a review's factual accuracy, reasoning depth, and consistency in ratings.

% \yz{I believe there are some other related work for LLM-based paper reviewing. Please refer to the 1st and 2nd paragraphs in the Introduction. My suggestion is to shorten the ``Agentic System'' paragraph and give more room to add these papers into the ``LLMs for Paper Review'' paragraph.}

\section{\benchmark}
% \label{sec:methods}
\label{sec:reviewbench}
% \zf{mention James Zou review refine paper}
% \zf{Note that 当我们评估 candidate review 的时候 评估 candidate reviewer models never observe $r^{*}_p$ or $\mathsf{R}^{\text{paper}}_p$ at generation time. Implementation details and prompts are provided in Appendix~\todo{ref}.}

\begin{figure}[t]
    \centering
    \includegraphics[width=\linewidth]{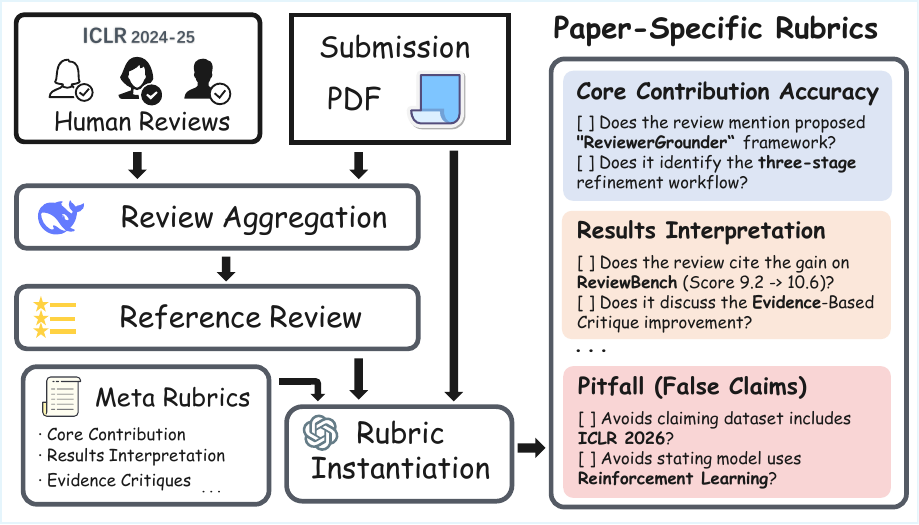}
    \caption{Overview of the \benchmark construction pipeline. For each paper, paper-specific rubrics are instantiated by an aggregated reference review, the submission PDF, and meta-rubrics.}
    \label{fig:benchmark}
    \vspace{-1em}
\end{figure}

Similarity-based metrics and LLM-as-a-Judge approaches used by prior studies for evaluating LLM-based reviewers~\cite{tan2024peer,zhu2025deepreview,zeng2025reviewrl} either fail to capture fine-grained review competencies or rely on ambiguous evaluation criteria and exhibit limited alignment with human judgments. To address these issues, we introduce \textbf{\benchmark}, a benchmark built on DeepReview-13K~\citep{zhu2025deepreview} that augments each paper $p$ and its human reviews $\mathsf{H}_p$ with two derived artifacts: (1) an aggregated reference review $r^{*}_p$; and (2) a set of paper-specific rubrics $\mathsf{R}^{\text{paper}}_p$. By leveraging the reference review $r^{*}_p$ and customized rubrics $\mathsf{R}^{\text{paper}}_p$ alongside an evaluator $\mathcal{E}$, \benchmark enables accurate, multi-faceted, and human-aligned assessment of LLM-generated reviews. The overview of \benchmark is illustrated in Figure~\ref{fig:benchmark}. We introduce the details of dataset construction in Section~\ref{sec:review_data}, followed by the description of the evaluation approach in Section~\ref{sec:rubric_eval} and Section~\ref{sec:numeric_eval}. Implementation details are provided in \S\ref{app:rubrics} and \S\ref{app:prompt_reviewbench}.

\subsection{Dataset Construction}
\label{sec:review_data}
\benchmark is constructed from DeepReview-13K~\citep{zhu2025deepreview}, which 
contains ICLR submissions and reviews from 2024 to 2025.\footnote{Since ICLR 2026 decisions have not yet been released, we restrict our data source to DeepReview-13K.}

\smallskip
\noindent \textbf{Data Filtering.}
We retain only papers with non-empty PDF-to-text content.
Specifically, we exclude (1) empty or incomplete submissions; (2) desk-rejected or withdrawn papers; (3) papers with fewer than three complete human reviews; and (4) papers missing mandatory review fields required for normalization, including textual sections and numeric scores (described below).
After filtering, we obtain a curated pool of approximately 12K papers.
Following prior work~\citep{zhu2025deepreview, zeng2025reviewrl}, we sample $N\approx$ 1.3K papers (about $10\%$ of the dataset) from this pool using a fixed random seed of 42.

\smallskip
\noindent \textbf{Human Review Normalization.}
For each paper $p$, we normalize its human reviews $\mathsf{H}_p$ into a unified schema aligned with the official ICLR review template~\citep{iclr_reviewer_guide_2026}. Each review is represented by (1) \emph{Textual Assessments}, including \textsc{Summary}, \textsc{Strengths}, \textsc{Weaknesses}, and \textsc{Questions}; and (2) \emph{Scores and Decisions}, comprising an overall rating in $[1,10]$, categorical scores (Soundness, Presentation, Contribution, and Confidence) in $[1,5]$, and the final decision (Accept/Reject). This filtering ensures all retained papers can be mapped to a consistent schema without per-paper exception handling.

\smallskip
\noindent \textbf{Reference Review Aggregation.}
For each paper $p$, we construct an aggregated reference review $r^{*}_p$ by consolidating the textual content of its human-written reviews $\mathsf{H}_p$ using DeepSeek-R1-Distill-Qwen-32B~\citep{guo2025deepseek}. We define the ground-truth rating $s_p$ as the mean of the human overall ratings in $\mathsf{H}_p$ and obtain the ground-truth decision $d_p$ from the dataset metadata. To ensure structural completeness, we populate the rating and decision fields of $r^{*}_p$ with $s_p$ and $d_p$, respectively.

% We follow DeepReview's default summarization settings and only modify the prompt to additionally condition on the paper text.
\subsection{Rubric-based Evaluation}
\label{sec:rubric_eval}
% 1. 基于 meta-rubrics 为它们构造 paper-specific rubrics 
% 2. 用这个 rubrics 来评估 candidate review 
The pipeline consists of three components: (1) a set of paper-agnostic meta-rubrics $\mathsf{R}^{\text{meta}}$ that define the multi-faceted criteria for high-quality reviews; (2) paper-specific rubrics $\mathsf{R}^{\text{paper}}_p$, instantiated from $\mathsf{R}^{\text{meta}}$ using the reference review $r^{*}_p$ to enable fine-grained, concrete evaluation of a candidate review $\hat{r}_p$; and (3) a fixed evaluator $\mathcal{E}$ that applies these rubrics to model-generated review $\hat{r}_p$ on a discrete ordinal scale to produce the final evaluation scores.

\subsubsection{Meta-Rubrics}
\label{sec:meta_rubric}

We define eight paper-agnostic meta-rubrics $\mathsf{R}^{\text{meta}}$ $=\{\mathsf{R}^{\text{meta}}_1,\ldots,\mathsf{R}^{\text{meta}}_8\}$, each capturing a distinct dimension of review quality.
This rubric set is derived from established peer-review standards, including reviewer guidelines from ICML, ICLR, and NeurIPS~\citep{icml_reviewer_instructions_2026, iclr_reviewer_guide_2026, neurips_reviewer_guidelines_2025}, and is iteratively refined with expert human feedback to ensure comprehensive coverage and operational clarity.
The eight dimensions are:
(1) \textsc{Core Contribution Accuracy},
(2) \textsc{Results Interpretation},
(3) \textsc{Comparative Analysis},
(4) \textsc{Evidence-Based Critique},
(5) \textsc{Critique Clarity},
(6) \textsc{Completeness Coverage},
(7) \textsc{Constructive Tone}, and
(8) \textsc{False or Contradictory Claims} (pitfall).
Each meta-rubric specifies (1) a polarity (positive vs.\ negative pitfall), (2) a concise checklist of key points, and (3) a scoring rule employed by the scoring model.
Full meta rubric definitions are provided in \S\ref{app:meta_rubrics}.

\subsubsection{Paper-specific Rubrics Construction}
\label{sec:paper_rubric}

Meta-rubrics $\mathsf{R}^{\text{meta}}$ define the \emph{general dimensions} that decide a high-quality review without elaborating customized rubrics for a specific instance. Therefore, we instantiate each meta-rubric $\mathsf{R}^{\text{meta}}_i$ using reference review $r_p^*$ into a paper-specific rubric $\mathsf{R}^{\text{paper}}_{p,i}$ and obtain
\begin{equation}
\mathsf{R}^{\text{paper}}_p=\{\mathsf{R}^{\text{paper}}_{p,i}\}_{i=1}^{8}.
\end{equation}
Each $\mathsf{R}^{\text{paper}}_{p,i}$ is a concise checklist of concrete, verifiable requirements grounded in the context of the paper, such as key claims, main results, or the most relevant comparisons for that work.

% how the paper-specific rubrics are instantiated
\smallskip
\noindent \textbf{Instantiation Procedure.}
We generate the paper-specific rubric set $\mathsf{R}^{\text{paper}}_p$ using a fixed rubric instantiation model based on GPT-OSS-120B~\citep{agarwal2025gpt}, conditioned on three inputs:
% \yw{Not sure but do we want to explain why not using GPT-4o since it has already been used in REVIEWGROUNDER?}\zf{We also use GPT-OSS-120B in \method}
 the paper text $p$, the meta-rubric set $\mathsf{R}^{\text{meta}}$, and the aggregated reference review $r^{*}_p$ (Sec.~\ref{sec:review_data}). To avoid human-phrase leakage, the reference review is used solely to ensure coverage of key issues raised by reviewers, not as a stylistic template. Consequently, each rubric item must be (1) grounded in verifiable evidence from the paper (e.g., a claim, section, table/figure, or comparison) and (2) independently checkable rather than copied from $r^{*}_p$.

% To reduce human-phrase leakage,  The reference review serves only as a coverage prior to highlighting salient issues raised by human reviewers, rather than being treated as a stylistic target to approximate. , we require each rubric item to be (i) grounded in verifiable paper evidence (e.g., a specific claim, section, table/figure, or comparison) and (ii) stated in an independently checkable form rather than copied from $r^{*}_p$.

\subsubsection{Scoring and Aggregation}
\label{sec:rubric_scoring}

Given a paper instance $p$, a candidate review $\hat{r}_p$, and the pre-generated customized rubrics $\mathsf{R}^{\text{paper}}_p$, we use GPT-OSS-120B as an LLM-evaluator to assign a discrete score for each rubric dimension:
\begin{equation}
s_{p,i}=\mathrm{Eval}\!\left(p,\hat{r}_p,\mathsf{R}^{\text{paper}}_{p,i}\right), \ \ i\in\{1,\ldots,8\}.
\end{equation}

The rubric set includes seven positive dimensions and one negative pitfall dimension. For positive dimensions, we use an ordinal scale ${0,1,2}$ corresponding to not satisfied, partially satisfied, and fully satisfied. For the negative pitfall dimension (i.e., \textsc{False or Contradictory Claims}), we use ${-2,-1,0}$ representing severe, mild, and none. Detailed scoring rules are provided in \S\ref{box:scoring_rules}.

\smallskip
\noindent \textbf{Overall Score.}
We define the overall content score as the sum of the individual dimension scores:
\begin{equation}
S(p,\hat{r}_p) = \sum_{i=1}^{8} s_{p,i}.
\end{equation}
We also provide the 8-dimensional score vector $(s_{p,1},\ldots,s_{p,8})$ for diagnostic analysis.

% \algReviewBench{}

\begin{figure*}[t]
    \centering
    \includegraphics[width=\textwidth]{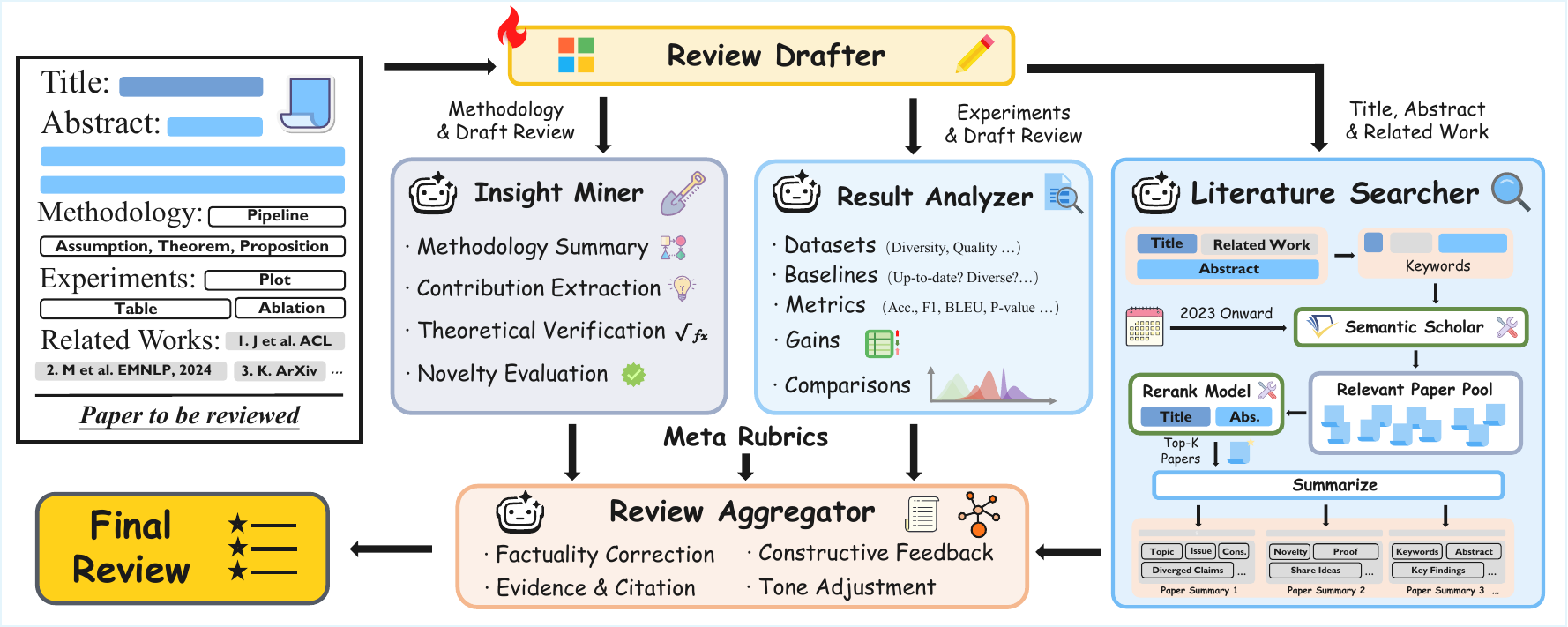}
    \caption{Overview of the \method. \method decomposing reviewing into collaborating agents:
    (a) \textbf{Review Drafter}: Generates an initial draft based on the paper. (b) Multi-dimensional Grounding Agents. \textbf{Literature Searcher}: Retrieves and summarizes related work using external tools. \textbf{Insight Miner}: Verifies methodology and core contributions. \textbf{Result Analyzer}: Checks experimental results. (c) \textbf{Review Aggregator}: Synthesizes the draft and evidence into a coherent, accurate, and actionable review.}
    \vspace{-1em}
    \label{fig:reviewgrounder} 
\end{figure*}

\subsection{Numeric-Field Evaluation}
\label{sec:numeric_eval}
Besides our proposed rubrics-based evaluation above, we also adopt a numeric-field-based approach to evaluate the numerical rating and final decision in a candidate review $\hat{r}_p$. Each rating is compared to the ground-truth $s_p$ (Sec.~\ref{sec:review_data}) using Mean Squared Error (MSE) and Mean Absolute Error (MAE), while the decision field is evaluated against the ground-truth $d_p$ (Sec.~\ref{sec:review_data}) using Accuracy (ACC) and F1 score.

% ReviewGrounder
\section{\method}
\label{sec:review_agent}

% \zf{Motivation should be clearer: 1. proprietary LLM (GPT, Claude) can not follow format and conference style instruction under long-context 2. Small open-source LLMs are generally less capable than large-scale proprietary models like GPT e.g, semantic understanding, reasoning, and comprehensive evaluation.}

We now introduce \method, a framework for producing grounded, substantive reviews.
It casts reviewing as a staged process that progressively refines an initial draft via targeted analysis, external evidence, and structured synthesis. An overview of \method is demonstrated in Figure~\ref{fig:reviewgrounder}. 
Below, we detail the stages and the functions of each agent, with implementation details provided in \S\ref{app:prompt_reviewgrounder}.

\subsection{Stage I: Draft Review Generation}
\label{sec:agent_draft}

Given a paper $p$, the first step is to produce a coherent and well-organized draft review that captures the basic structure and stylistic conventions commonly observed in human-written reviews.
To this end, we fine-tune a pre-trained LLM on a subset of DeepReview-13K \citep{zhu2025deepreview} to obtain a \textbf{Drafter} $\mathcal{P}$.
Note that the training data is fully disjoint from \benchmark, and $\mathcal{P}$ primarily acquires shallow review patterns, such as section-level organization and syntax-level cues, rather than grounded reasoning.
Given $p$, the Drafter $\mathcal{P}$ then generates an initial draft review $r^{(0)}$.

\subsection{Stage II: Multi-dimensional Review Grounding}
\label{sec:agent_tools}

The initial draft $r^{(0)}$ often lacks comparative context and evidential support.
To address these gaps, we introduce three specialized agents that collaboratively enrich and consolidate the draft along key rubric-defined review dimensions.
Each agent operationalizes a core reviewing capability that cannot be reliably captured by single-pass drafting alone.
Specifically, the \textbf{Literature Searcher} $\mathcal{S}$ situates the submission within contemporary literature to support grounded novelty assessment;
the \textbf{Insight Miner} $\mathcal{M}$ consolidates conceptual understanding by analyzing the paper's core contributions and technical claims;
and the \textbf{Result Analyzer} $\mathcal{A}$ strengthens empirical grounding by examining experimental design, results, and quantitative evidence.
Together, these agents operate in parallel to iteratively refine $r^{(0)}$, producing an enriched review representation $\mathsf{E}(p)$ that enhances review substantiveness.
 
\smallskip
\noindent \textbf{Literature Searcher $\mathcal{S}$.} Given the title, abstract, and related work of the submission $p$, the Literature Searcher $\mathcal{S}$ operates as a tool-integrated agent. 
It first extracts a set of representative keywords that capture the paper's technical scope, and uses them to query the Semantic Scholar API~\cite{kinney2023semantic,bragg2025astabench}, retrieving candidate papers published from 2023 onward.
Retrieved papers are fed into an off-the-shelf reranker\footnote{\url{https://huggingface.co/OpenSciLM/OpenScholar_Reranker}}, and the top-10 most relevant works are selected.

For each selected paper, $\mathcal{S}$ produces a concise, structured debrief summarizing its core methodology, main findings, and experimental evidence most relevant to $p$.
This process explicitly grounds comparative analysis and supports informed assessment of $p$'s novelty and positioning.

% actually: 
% 1. get topK=10 papers from all keywords-related papers
% 2. for each paper, summarize its content, main contribution and its correspondence with the target paper (related-work debrief)
% 3. organize all 10 papers' debriefs into a structuralized json object 
% Finally, $\mathcal{S}$ merges papers across all keywords to generate a comprehensive, up-to-date related work for $p$.

% 1. combine rubrics how to evaluation 2. deepreview 
\smallskip
\noindent \textbf{Insight Miner $\mathcal{M}$.} The Insight Miner $\mathcal{M}$ targets the conceptual and methodological core of $p$.
It retrieves sections relevant to the technical approach, distills the central contributions, and evaluates the validity of the paper's novelty claims and stated differences.
Based on this analysis, $\mathcal{M}$ refines the method-focused parts of the draft review $r^{(0)}$ by providing actionable suggestions grounded in specific parts of the paper (e.g., sections or formulas). This process enhances the accuracy and substantiveness of discussions around model design, algorithmic formulation, optimization, and implementation, transforming vague or generic claims into precise, evidence-supported critiques.

\smallskip
\noindent \textbf{Result Analyzer $\mathcal{A}$.} Complementary to $\mathcal{M}$, the Result Analyzer $\mathcal{A}$ focuses exclusively on empirical evaluation.
It extracts key experimental elements, including datasets, baselines, evaluation metrics, performance gains, and statistical comparisons.
Using such signals, $\mathcal{A}$ refines the experiment-related components of $r^{(0)}$, ensuring that claims about performance and effectiveness are faithful to the reported results and grounded in concrete tables, figures, and quantitative comparisons.

\subsection{Stage III: Rubric-Guided Review Synthesis}
\label{sec:agent_refine}
% motivation 
% 1. accurate 2. 有见解的 actionable 4. comprehensive 

In the final stage, an Aggregator $\mathcal{G}$ synthesizes the outputs of all upstream agents to produce a coherent, accurate, and actionable review.
Specifically, $\mathcal{G}$ takes as input the paper $p$, the initial draft $r^{(0)}$, the grounded review representation $\mathsf{E}(p)$, and the generic meta-rubrics $\mathsf{R}^{\text{meta}}$, and performs a final round of consolidation and refinement.
Note that paper-specific rubrics $\mathsf{R}_p^{\text{paper}}$ used in \benchmark are not exposed at generation time, preventing evaluation leakage.

The Aggregator $\mathcal{G}$ is designed to:
(1) correct factual errors and ensure faithful characterization of the paper's contributions and methodology;
(2) strengthen critiques by anchoring them to specific parts of the paper, enabling grounded results interpretation, evidence-based critique, and comparative analysis;
(3) translate grounded observations into clear, constructive, and actionable suggestions;
and (4) produce a balanced and comprehensive assessment that aligns with rubric-level expectations on coverage, clarity, and tone.

\section{Experiments}
% \yw{Currently, the benchmark itself is not evaluated. It is unclear whether reviewers will raise concerns about this aspect. One possible approach that might be considered next time could be to assess the benchmark by examining whether reviews that achieve higher scores are better aligned with the reviews with high ratings/qualities. However, seems that currently we do not have such labeled data.}
% \zf{delete act rubrics}
% experiments: main and ablation
% main experiment: run our framework on the base 1300 samples (benchmark) against other sota models including agenticreview, deepreviewer, etc. 

% overall setting of the experiment: what do we evaluate and what are the metrics
\noindent \textbf{Experimental Setup.}
We conduct evaluation on \benchmark (Sec.~\ref{sec:reviewbench}) using two complementary families of metrics:
(1) \textit{Rubric-based Evaluation}, which assesses the textual quality of generated reviews across eight paper-specific rubric dimensions (Sec.~\ref{sec:rubric_eval}); and
(2) \textit{Numeric-field Evaluation}, which measures predicted ratings with MSE/MAE and decisions with ACC/F1 (Sec.~\ref{sec:numeric_eval}). To ensure fair comparison and prevent protocol leakage, we fix the paper-specific rubrics $\mathsf{R}^{\text{paper}}_p$ and the evaluator $\mathcal{E}$ across all methods. During review generation, models access only the paper text $p$ and are never exposed to the aggregated reference review $r^{*}_p$ or the paper-specific rubrics $\mathsf{R}^{\text{paper}}_p$.

% evaluation baselines: categories of the baselines that we are comparing our framework against
\smallskip
\noindent \textbf{Baselines.}
We compare \method against three categories of baselines: (1)\textit{ Foundation Models}, including Qwen3-32B~\citep{yang2025qwen3}, QWQ-32B~\citep{qwq32b}, GPT-4o~\citep{hurst2024gpt}, and GPT-4.1~\citep{openai_gpt41_2025}; (2) \textit{Agentic Reviewing Framework}, including AI Scientist~\citep{lu2024ai}, AgentReview~\citep{jin2024agentreview}, which are instantiated with GPT-4o and GPT-4.1 as backbone models; and (2) \textit{Fine-tuned Reviewer Models}, including CycleReviewer-8B/70B~\citep{weng2024cycleresearcher} and DeepReviewer-7B/14B~\citep{zhu2025deepreview}. More details on baseline implementations are in \S\ref{app:baseline}. 

\smallskip
\noindent \textbf{Implementation Details.}
In our main experiments, \textit{Drafter} is instantiated with Phi-4-14B~\citep{abdin2024phi}, while the remaining modules (i.e., the \textit{Literature Searcher}, \textit{Insight Miner}, \textit{Result Analyzer}, and \textit{Aggregator}) are instantiated with GPT-OSS-120B~\citep{agarwal2025gpt}. Among these, only the \textit{Drafter} is trainable. We train the model using a portion of the DeepReview-13K dataset. Further implementation details on \method can be found in \S\ref{app:implementation_details}.
% 1. Reviewgrounder 中 agent: drafter -> phi4-14B (SFT details)  / Insight Miner / Result Analyzer is an instance of GPT-OSS-120B

% 2. review generation temperature 

% \smallskip
% \noindent \textbf{Implementation Details.}
% \label{sec:agent_impl}
% We use the paper text provided by DeepReview-13K, where each submission is converted from PDF into plain text.
% Thanks to \method{}'s modular, plug-and-play design, we experiment with multiple backbone reviewer models $M_{\text{rev}}$ (DeepReviewer-7B/14B and an in-house model).
% The in-house model is based on Qwen3-4B and is trained on the DeepReview-13K training split.
% The aggregated reference review model is fixed to DeepReview's DeepSeek-R1-Distill-Qwen-32B.

% All other LLM components in our benchmark and ReviewerAgent, including rubric instantiation, rubric scoring, keyword generation, related-work retrieval and summarization, results summarization, and refinement, are supported by GPT-OSS-120B.
% Unless otherwise stated, we run GPT-OSS-120B with vLLM default sampling parameters and set the generation limit to $37{,}000$ tokens.
% For rubric scoring, we use deterministic decoding (temperature $0$) to improve reproducibility. \todo{list any additional fixed decoding parameters if needed}

\subsection{Main Experiment Results}

\subsubsection{Rubric Evaluation Results}
% rubric table 
\begin{table*} \vspace{-1mm}
    \centering
    \footnotesize
    \setlength{\tabcolsep}{0.5em}
    \renewcommand{\arraystretch}{0.95}
    \resizebox{\textwidth}{!}{
\begin{tabular}{@{}cccccccccc|cc@{}}
\toprule
\textbf{Method} &
  \multicolumn{1}{c|}{\textbf{Model}} &
  \textbf{Core} &
  \textbf{Res.} &
  \textbf{Comp.} &
  \textbf{EBC} &
  \textbf{Clr.} &
  \textbf{Cov.} &
  \textbf{Tone} &
  \textbf{Contradict.} &
  \textbf{Overall} &
  \textbf{$\Delta$} \\ \midrule
\multirow{4}{*}{Foundation Model} &
  \multicolumn{1}{c|}{Qwen3-32B} &
  1.6971 & 0.7642 & 0.5800 & 0.1437 &
  1.6128 & 1.1537 & 1.9992 & -0.1460 &
  7.8047 & \mygreen{38}{\gain{38\%}} \\
 &
  \multicolumn{1}{c|}{QWQ-32B} &
  1.6901 & 0.6531 & 0.3513 & 0.1186 &
  1.6792 & 0.9461 & 1.9969 & -0.0836 &
  7.3517 & \mygreen{46}{\gain{46\%}} \\
 &
  \multicolumn{1}{c|}{GPT-4o} &
  1.1969 & 0.1037 & 0.0302 & 0.0024 &
  1.0499 & 0.3318 & 1.9840 & -0.1233 &
  4.5756 & \mygreen{100}{\gain{135\%}} \\
 &
  \multicolumn{1}{c|}{GPT-4.1} &
  1.7573 & 0.6966 & 0.3406 & 0.1074 &
  1.6327 & 1.1675 & 1.9992 & -0.0397 &
  7.6616 & \mygreen{41}{\gain{41\%}} \\ \midrule
\multirow{2}{*}{AgentReview} &
  \multicolumn{1}{c|}{GPT-4o} &
  1.1300 & 0.1600 & 0.1100 & 0.1250 &
  1.3400 & 0.5900 & 2.0000 & -0.1600 &
  4.8675 & \mygreen{100}{\gain{121\%}} \\
 &
  \multicolumn{1}{c|}{GPT-4.1} &
  1.0300 & 0.1300 & 0.1200 & 0.0000 &
  1.4100 & 0.6300 & 1.9800 & -0.1600 &
  4.9620 & \mygreen{100}{\gain{117\%}} \\ \midrule
\multirow{2}{*}{AI Scientist} &
  \multicolumn{1}{c|}{GPT-4o} &
  0.8500 & 0.0000 & 0.0200 & 0.0000 &
  0.6700 & 0.1800 & 1.7600 & -0.1900 &
  3.6800 & \mygreen{100}{\gain{193\%}} \\
 &
  \multicolumn{1}{c|}{GPT-4.1} &
  1.6700 & 0.4800 & 0.3600 & 0.0830 &
  1.5600 & 1.1300 & 1.9400 & -0.0900 &
  7.0893 & \mygreen{52}{\gain{52\%}} \\ \midrule
\multirow{2}{*}{CycleReviewer} &
  \multicolumn{1}{c|}{Llama-3.1-8B} &
  0.9852 & 0.1011 & 0.0645 & 0.0089 &
  0.5832 & 0.1493 & 1.6571 & -0.4504 &
  3.0989 & \mygreen{100}{\gain{248\%}} \\
 &
  \multicolumn{1}{c|}{Llama-3.1-70B} &
  1.0187 & 0.1633 & 0.0980 & 0.0109 &
  0.7698 & 0.2551 & 1.8476 & -0.6412 &
  3.5220 & \mygreen{100}{\gain{206\%}} \\ \midrule
\multirow{2}{*}{DeepReviewer} &
  \multicolumn{1}{c|}{Phi-4-7B} &
  1.4207 & 0.4545 & 0.3299 & 0.1311 &
  1.3743 & 1.0599 & 1.9432 & -0.3953 &
  6.3183 & \mygreen{56}{\gain{70\%}} \\
 &
  \multicolumn{1}{c|}{Phi-4-14B} &
  1.6306 & 0.6532 & 0.4977 & 0.3532 &
  1.6772 & 1.2877 & 1.9930 & -0.1922 &
  7.9004 & \mygreen{25}{\gain{36\%}} \\ \midrule \rowcolor[HTML]{FFCCC9}
\textbf{\method{}} &
  \multicolumn{1}{c|}{\textbf{Phi-4-14B}} &
  \textbf{1.8507} & \textbf{1.4075} & \textbf{0.9059} & \textbf{1.4831} &
  \textbf{1.9191} & \textbf{1.3289} & \textbf{1.9992} & \textbf{-0.1245} &
  \textbf{10.7699} & - \\ \bottomrule
\end{tabular}
}
\caption{\textbf{Performance comparison of reviewer models on \benchmark{} under rubric-based evaluation.} We visualize gains of \method{} to each baseline in the \colorbox{DeltaBg}{$\Delta$ columns}. Notes: Higher scores indicate better performance; Contradict. is a pitfall dimension scored in ${-2,-1,0}$, 
while others are scored in ${0,1,2}$. \textit{Abbreviations:} Core=\textsc{Core Contribution Accuracy}, 
Res.=\textsc{Results Interpretation}, 
Comp.=\textsc{Comparative Analysis}, 
EBC=\textsc{Evidence-Based Critique}, 
Clr.=\textsc{Critique Clarity}, 
Cov.=\textsc{Completeness Coverage}, 
Tone=\textsc{Constructive Tone}, 
Contradict.=\textsc{False or Contradictory Claims}.}
\label{tab:main_rubric}
\end{table*}

Our main rubric-based content evaluation results are presented in Table~\ref{tab:main_rubric}. Overall, \method consistently outperforms all baseline models by substantial margins across every rubric dimension. Relative to the best-performing foundation model, Qwen3-32B, \method achieves an improvement of 38\%. 
Against specialized agentic frameworks, our \method surpasses AgentReview and AI Scientist (both based on GPT-4o) by 121\% and 193\%, respectively. Among fine-tuned models, \method exceeds the top performer, DeepReviewer-14B, by 36\%. Remarkably, \method also outperforms the $\sim$200B-parameter GPT-4o across all dimensions, with a gain of 135\%. A detailed analysis is provided in~\S\ref{app:main_results_analysis}.

\subsubsection{Numeric-Field Evaluation Results}
We additionally report numerical rating and final-decision evaluations in Table~\ref{tab:main_auto} for completeness and comparability with prior work~\citep{zhu2025deepreview}. Compared with all baselines, \method achieves the lowest rating error (MSE: 1.1607, MAE: 0.8597) and the highest decision prediction accuracy (ACC: 0.6809, F1: 0.6699). Relative to the strongest AI Scientist variant (Gemini-2.0-Flash-Thinking), \method improves ACC by 8\% while reducing MSE by approximately 63\%. Even when compared with strong fine-tuned models such as DeepReviewer-14B, \method consistently improves both rating accuracy and decision quality, demonstrating its ability to produce more reliable and consistent paper assessments.

% \method{} refines the \emph{textual} sections of a review, while leaving numeric fields (overall rating and categorical scores) unchanged from the backbone reviewer when they are present.\footnote{When a baseline does not output a given numeric field, we report ``--'' in \autoref{tab:main_auto}.}
% Accordingly, numeric-field performance primarily reflects the backbone model rather than the refinement procedure, whereas our main improvements are consistently observed in rubric-based content evaluation (Table~\ref{tab:main_rubric}).

% This behavior is visible in \autoref{tab:main_auto}: with DeepReviewer-14B as the backbone, the numeric-field metrics remain identical before and after applying \method{} (e.g., MSE-Rating $1.3340$, MAE-Rating $0.8998$, Spearman-Rating $0.4320$, and decision accuracy $0.6798$). In contrast, swapping the backbone changes these metrics accordingly: \method{} with a Qwen3-4B backbone yields MSE-Rating $2.2801$ and decision accuracy $0.5480$, reflecting backbone-dependent calibration on numeric prediction.

% Overall, these results suggest that improvements in review-text quality (Table~\ref{tab:main_rubric}) can be largely decoupled from accurate score/decision prediction. Extending the framework to explicitly calibrate numeric fields is a natural direction for future work.

\subsection{Analysis of Performance Gains}
This section examines several factors influencing the performance of \method on \benchmark under rubric-based evaluation.

% numric table 
\begin{table} 
    \centering
    \footnotesize
    \setlength{\tabcolsep}{0.5em}
    \renewcommand{\arraystretch}{0.95}
    \resizebox{\linewidth}{!}{
    \begin{tabular}{@{}cc|cc|cc@{}}
    \toprule
    \multirow{2}{*}{\textbf{Method}} & \multirow{2}{*}{\textbf{Model}} & \multicolumn{2}{c|}{\textbf{Decision}} & \multicolumn{2}{c}{\textbf{Rating}} \\ \cmidrule(l){3-6}&
    & \textbf{ACC$\uparrow$} 
    & \textbf{F1$\uparrow$} 
    & \textbf{MSE$\downarrow$} 
    & \textbf{MAE$\downarrow$}
    \\ \midrule
    
    \multirow{3}{*}{AgentReview}    
    & Claude-3-5-sonnet               
    & 0.2826 
    & 0.2541 
    & 2.8406 
    & 1.2989
    \\
    & Gemini-2.0-Flash-Thinking 
    & 0.4242       
    & 0.4242      
    & 2.6186       
    & 1.2170       
    \\
    & DeepSeek-V3
    & 0.3140
    & 0.2506
    & 1.9951
    & 1.1017
    \\ \midrule
    
    \multirow{5}{*}{AI Scientist} 
    & GPT-o1
    & 0.4167
    & 0.4157
    & 4.3072
    & 1.7917
    \\
    & Claude-3-5-sonnet
    & 0.5579
    & 0.4440
    & 3.0992
    & 1.3500
    \\
    & Gemini-2.0-Flash-Thinking 
    & 0.6139
    & 0.4808
    & 3.9232
    & 1.6470
    \\
    & DeepSeek-V3
    & 0.4059
    & 0.3988
    & 4.8006
    & 1.8403
    \\
    & DeepSeek-R1
    & 0.4259
    & 0.4161
    & 4.7719
    & 1.8099
    \\ \midrule
    
    \multirow{2}{*}{CycleReviewer}   
    & Llama-3.1-8B
    & 0.2354	
    & 0.3988
    & 3.1324
    & 1.3663
    \\
    & Llama-3.1-70B                   
    & 0.1545	
    & 0.4156	
    & 1.8440	
    & 1.0643
    \\ \midrule
    
    \multirow{2}{*}{DeepReviewer}    
    & Phi-4-7B                        
    & 0.6381	
    & 0.6068	
    & 1.4442	
    & 0.9416
    \\
    & Phi-4-14B
    & 0.6667
    & 0.5204
    & 1.3527
    & 0.9041
    \\ \midrule 
    
    \rowcolor[HTML]{FFCCC9}
    \textbf{\method{}}          
    & \textbf{Phi-4-14B}
    & \textbf{0.6939}
    & \textbf{0.6699}
    & \textbf{1.1607}
    & \textbf{0.8597}
    \\ \bottomrule    
    \end{tabular}}
\caption{\textbf{Performance comparison of reviewer models on \benchmark under numeric-field evaluation.}  }
\label{tab:main_auto}
\end{table}

% ablation experiments: 
% ablation on what parameters? 
% related work tool:
% 1. form of paper summarization: 
%     - summarize each paper structurally OR
%     - summarize all papers into one sentence OR 
%     - directly give each paper's abstract OR
% 2. Asta's TopK (number of keywords and number of total papers), 
% 
% 3. Disable the related work search & summarization module
% experiment&result summarization tool:
% 1. Disable the experiment&result summarization module
% 2. only summarize the experiment 
% 3. only summarize the result

%
\subsubsection{Component Ablation}
We conduct a comprehensive ablation study to assess the contribution of each agent in \method, as summarized in Table~\ref{tab:ablation_agent}. Detailed analyses are provided in \S\ref{app:ablation_study}.

% 使用相同数据 tune SFT-tuned Qwen3-4B 比 Phi-4-7B 效果好但是 比不过 14B 这说明

\smallskip
\noindent \textbf{Impact of Drafter Backbones.} As shown in Table~\ref{tab:ablation_agent}, when trained on the same SFT data, Qwen3-4B consistently outperforms Phi-4-7B but remains inferior to Phi-4-14B, indicating that model capacity continues to play a critical role in draft review synthesis quality. Nevertheless, results with smaller Drafters (Phi-4-7B and Qwen3-4B) demonstrate that the framework remains effective (e.g., Qwen3-4B: 10.6418), suggesting that the proposed grounding and aggregation mechanisms substantially benefit smaller models, even though stronger backbones achieve superior overall performance.

\begin{table}
\setlength{\tabcolsep}{0.5em}
\renewcommand{\arraystretch}{0.9}
\resizebox{\linewidth}{!}{
\begin{tabular}{ccc||ccc|c}
\toprule
\multirow{2}{*}{\textbf{\makecell{Drafter\\Phi-4-14B}}} &
\multirow{2}{*}{\textbf{\makecell{Drafter\\Phi-4-7B}}} &
\multirow{2}{*}{\textbf{\makecell{Drafter\\Qwen3-4B}}} &
\multirow{2}{*}{\textbf{Searcher}} &
\multirow{2}{*}{\textbf{Miner}} &
\multirow{2}{*}{\textbf{Analyzer}} &
\multirow{2}{*}{\textbf{Overall}} \\
 & & & & & & \\
\midrule
 & & \checkmark & \checkmark & \checkmark & \checkmark & 10.6418 \\
 & \checkmark &  & \checkmark & \checkmark & \checkmark & 10.5928 \\
\checkmark & &  &  & \checkmark & \checkmark & 10.6568 \\
\checkmark & &  & \checkmark &  & \checkmark & 10.6526 \\
\checkmark & &  & \checkmark & \checkmark &  & 10.0186 \\
\midrule
\checkmark & &  & \checkmark & \checkmark & \checkmark & \textbf{10.7699} \\
\bottomrule
\end{tabular}
}
\caption{\textbf{Ablation Study of \method under rubric-based evaluation.}}
\label{tab:ablation_agent}
\vspace{-1em}
\end{table}

\smallskip
\noindent \textbf{Impact of Grounding Agents.} We further evaluate the contribution of the three proposed grounding agents, the Literature Searcher $\mathcal{S}$, the Insight Miner $\mathcal{M}$, and the Result Analyzer $\mathcal{A}$, by individually removing each from the full pipeline, using Phi-4-14B as the Drafter. As shown in Table~\ref{tab:ablation_agent}, omitting any agent leads to a performance degradation relative to the full model (10.7699), underscoring the importance of each component.

\subsubsection{Hyperparameter Study}

Additionally, we analyze the sensitivity of the Literature Searcher $\mathcal{S}$.

\begin{figure}[t]
    \centering
    \includegraphics[width=\linewidth]{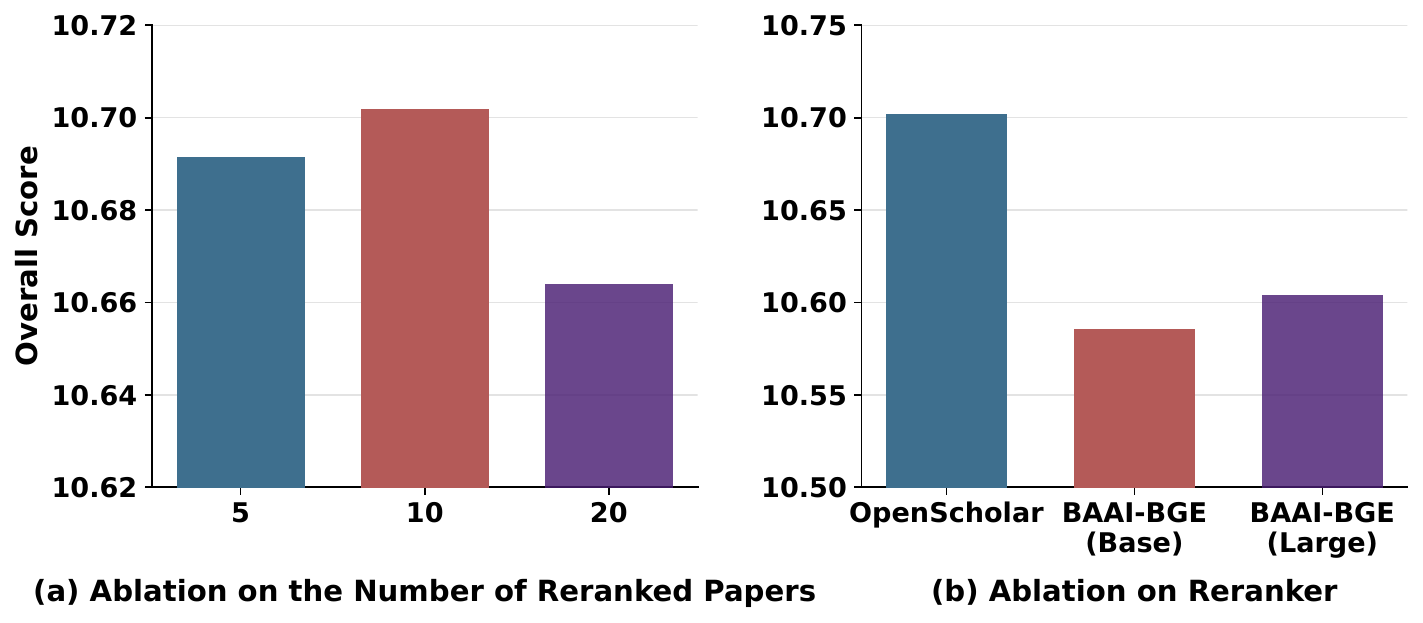}
    \caption{\textbf{Ablation on Literature Searcher configurations under rubric-based evaluation}.}
    % (a) Impact of the number of reranked papers ($N$) fed into the context. (b) Performance comparison between different underlying retriever models (OpenResearcher vs. BAAI-General) 
    \label{fig:ablation}
    % \vspace{-1em}
\end{figure}

\smallskip
\noindent \textbf{Number of Reranked Papers.} As shown in Figure~\ref{fig:ablation}(a), offering Literature Searcher $\mathcal{S}$ with 10 reranked papers per keyword (Sec.~\ref{sec:agent_tools}) yields the highest overall score. Using fewer papers (5) provides insufficient contextual information, whereas retrieving more papers (20) introduces additional noise that can distract the Drafter.

\smallskip
\noindent \textbf{Reranker Selection.} Figure~\ref{fig:ablation}(b) compares the impact of the reranking model used to reorder retrieved papers in $\mathcal{S}$. OpenScholar-Reranker~\citep{asai2024openscholar}, a variant of BAAI-BGE-Large~\citep{bge_embedding} fine-tuned for scientific literature synthesis, significantly outperforms BAAI-BGE in both its Base and Large versions. This result highlights the importance of domain-specific retrievers for accurately identifying relevant related work.

\subsection{Defend Attacks Analysis}
% 1. settings 2. figure 3. conclusion 
\begin{figure}[t]
    \centering
    \includegraphics[width=\linewidth]{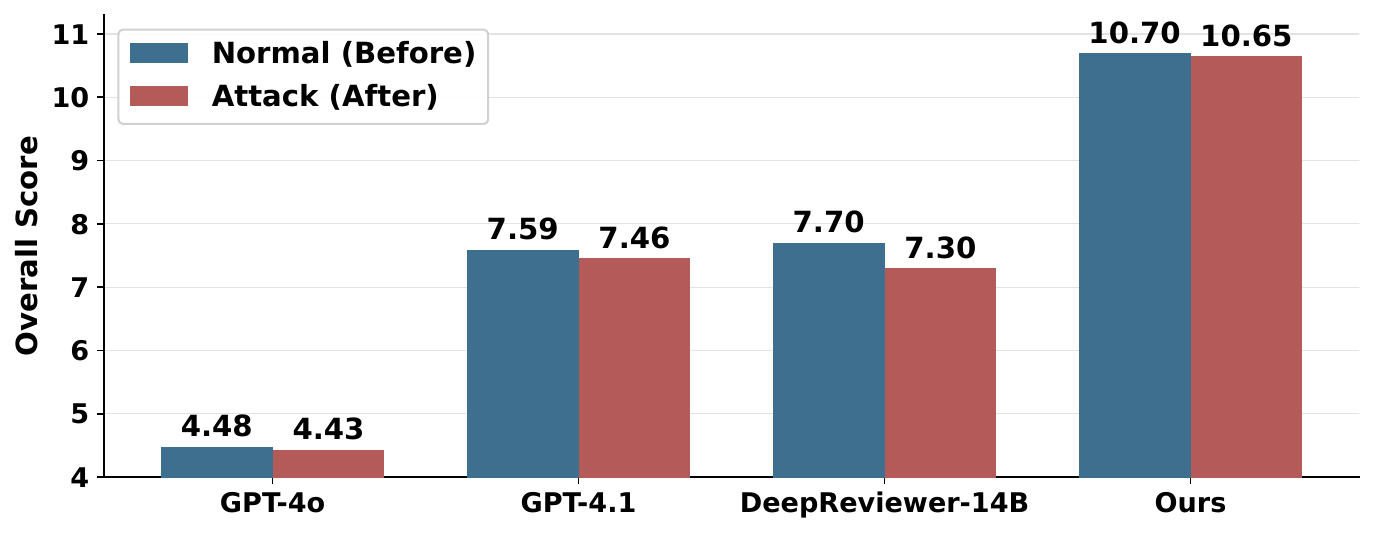}
    \caption{
        \textbf{Comparison with baselines under normal and
attack scenarios via rubric-based evaluation.} 
    }
    \label{fig:attack}
    \vspace{-1em}
\end{figure}

We evaluate the robustness of \method against adversarial attacks~\cite{ye2024we} by injecting malicious instructions into input papers. For evaluation, we randomly sample $500$ papers from \benchmark, with the rubric-based evaluation results presented in Figure~\ref{fig:attack}. We can find that (1) \textbf{Baseline vulnerability}: Existing reviewer models are significantly affected by the attack; for instance, DeepReviewer-14B drops from 7.70 to 7.30, demonstrating clear susceptibility to adversarial instructions. (2) \textbf{Robustness of \method}: In contrast, \method shows strong resilience. Despite a high baseline score, its performance remains largely stable, with only a minor decrease of 0.05 points (10.70 to 10.65). (3) \textbf{Value of \benchmark and rubric-based evaluation}: These results underscore the importance of \benchmark. Since rubric-based evaluation emphasizes semantic quality rather than absolute scores, malicious instructions can mislead review-generation models, causing them to ignore prior guidelines and produce lower-quality reviews, which in turn results in lower scores—thereby preventing authors from artificially inflating review scores by injecting instructions.

\subsection{Human Evaluation}
\begin{table}[t]
\small
\centering
\resizebox{\linewidth}{!}{
\begin{tabular}{cccc}
    \toprule
    \textbf{MAE} & \textbf{Spearman ($\rho$)} & \textbf{Pearson ($r$)} & \textbf{Pairwise Error} \\
    \midrule
    0.0969 & 0.7923 & 0.8954 & 0.1494 \\
    \bottomrule
\end{tabular}
}
\caption{
\textbf{Alignment metrics between experts and automated judgments.}
}
\vspace{-1em}
\label{tab:human_eval_overall}
\end{table}

To validate the effectiveness and robustness of rubric-based evaluation in \benchmark, we conducted a human study. We randomly sampled 120 papers from \benchmark and asked human experts, each with a strong publication record averaging 2,000 Google Scholar citations, to rate the reviews generated by \method according to the meta rubrics $\mathsf{R}^{\text{meta}}$ (Sec.~\ref{sec:meta_rubric}). The resulting human overall scores were then compared with those produced by the LLM-evaluator using paper-specific rubrics (Sec.~\ref{sec:rubric_scoring}), enabling us to assess alignment between expert judgments and automated evaluations. As shown in Table~\ref{tab:human_eval_overall}, human scores closely align with the LLM-evaluator, with a Pearson correlation of 0.8954 and a Spearman correlation of 0.7923. Error rates are also low, with a Mean Absolute Error of 0.0969 and a Pairwise Absolute Error of 0.1494, demonstrating that paper-specific rubrics provide an effective and robust approach for review evaluation.
\section{Conclusion}
We present \method, a rubric-guided, tool-integrated multi-agent framework that rethinks LLM-based peer review as a staged process of drafting and grounding.
\method explicitly decomposes review construction into complementary roles that retrieve relevant literature, analyze paper-specific evidence, and synthesize critiques guided by reviewer guidelines.
Together with \benchmark, a benchmark enabling multi-faceted, rubric-based, and human-aligned evaluation, our work provides an instance-specific evaluation lens and an effective modeling approach for LLM-based peer review.
Extensive experiments demonstrate consistent improvements across key review dimensions, with \method outperforming baselines with substantially larger/stronger backbones.
Additional analyses confirm the complementary role of each agent and the robustness of the framework under adversarial attacks.

\section*{Limitations}
While \method achieves strong performance on \benchmark, there are several limitations that motivate future work.
First, due to infrastructure constraints, we do not implement in-the-flow training of the multi-agent workflow~\citep{li2025flow}, which coordinates agent modules in a trainable, end-to-end manner.
Although our staged approach with separate drafter and grounding agents already attains state-of-the-art results, enabling in-the-flow learning may further improve inter-agent coordination and overall review quality.
Second, our study focuses exclusively on LLM-based reviewers and does not explore higher-level constructs such as LLM-based meta-reviewers or the potential for iterative feedback loops~\cite{feedback}.
Investigating how reviews and feedback could mutually enhance each other, for example, through multi-turn interaction, remains an interesting direction for future research.
Finally, due to differences in public availability and practical constraints across venues, the cross-venue set we curated is necessarily partial and uneven, limiting the completeness of our framework's performance and robustness evaluation. 

\section*{Ethical Considerations}
The development of \method and \benchmark carries important ethical considerations given their role in supporting scientific peer review.
While our framework aims to enhance review consistency, evidence grounding, and rubric adherence, it is not intended to replace human reviewers.
Over-reliance on LLM-generated reviews could risk diminishing critical evaluation skills, introducing unintended biases from training data, or amplifying systematic preferences toward certain topics or methodologies.
To mitigate these risks, we emphasize a human-in-the-loop approach: \method's outputs are designed to assist expert reviewers, who should critically assess, verify, and contextualize all generated feedback.
Moreover, we encourage careful auditing of model outputs, ongoing monitoring for bias, and iterative refinement of both models and evaluation criteria, aiming to support responsible and informed use of LLM-assisted peer review.

\end{spacing}

\bibliography{main}

\clearpage
\appendix
% \onecolumn

% Create a new ToC for appendix only
\section*{Appendix Contents}
\setcounter{tocdepth}{2}
\renewcommand{\contentsname}{Appendix Contents}
\startcontents[appendix]  % Requires the 'titletoc' package
\printcontents[appendix]{}{1}{}

\section{Additional Experiments}

\subsection{Generalization Beyond ICLR}

To evaluate the broader generalization of \method, we evaluated our framework under a subset of the combination of reviews from multiple venues with diverse topics, including NeurIPS, AAAI, ACM Multimedia, and CVPR (2023–2024) via both OpenReview’s official API and from an open-source public Huggingface dataset\footnote{\url{https://huggingface.co/datasets/guochenmeinian/openreview_raw}}. We randomly sampled approximately 1.7K papers for this experiment. The literature PDFs were converted into Markdown format using Nougat OCR\footnote{\url{https://facebookresearch.github.io/nougat/}}, followed by a custom cleaning pipeline involving regular expressions to remove OCR artifacts, redundant symbols, and formatting inconsistencies.

\begin{table*}[t]
\centering
\caption{\textbf{Cross-domain rubric-based evaluation on the expanded review dataset.} Higher scores indicate better performance. \textit{Contradict.} is a pitfall dimension scored in $\{-2,-1,0\}$, where higher is better (fewer false or contradictory claims), while all other dimensions are scored in $\{0,1,2\}$.}
\label{tab:appendix_domain_coverage}
\small
\setlength{\tabcolsep}{5pt}
\begin{tabular}{lccccccccc}
\toprule
Model & Core & Res. & Comp. & EBC & Clr. & Cov. & Tone & Contradict. & Overall \\
\midrule
Only Drafter (14B) & 1.4393 & 0.4934 & 0.4606 & 0.3599 & 1.5775 & 1.1887 & 1.9318 & -0.5075 & 6.9436 \\
ReviewGrounder & \textbf{1.8025} & \textbf{1.3650} & \textbf{0.7700} & \textbf{1.5017} & \textbf{1.7475} & \textbf{1.3300} & \textbf{1.9850} & \textbf{-0.1100} & \textbf{10.3917} \\
\bottomrule
\end{tabular}
\vspace{2mm}

\begin{minipage}{0.98\linewidth}
\footnotesize
\textbf{Abbreviations:}
Core = \textsc{Core Contribution Accuracy};
Res. = \textsc{Results Interpretation};
Comp. = \textsc{Comparative Analysis};
EBC = \textsc{Evidence-Based Critique};
Clr. = \textsc{Critique Clarity};
Cov. = \textsc{Completeness Coverage};
Tone = \textsc{Constructive Tone};
Contradict. = \textsc{False or Contradictory Claims}.
\end{minipage}
\end{table*}

As shown in Table~\ref{tab:appendix_domain_coverage}, \method maintains robust cross-domain performance across machine learning, computer vision, multimedia, and general artificial intelligence venues, achieving an overall rubric score of 10.3917. Moreover, \method substantially outperforms the 14B drafter baseline, achieving a 49.6\% higher overall score (10.3917 vs. 6.9436), confirming the effectiveness of review-grounded refinement.

\subsection{Robustness Under Challenging Adversarial Attacks}

To further evaluate~\method's robustness under adversarial attacks, we conducted additional studies with three types of hidden perturbations: \emph{disguised misleading information}, \emph{niche terminology induction}, and \emph{scattered key information}. For this study, we randomly selected 100 papers from \textsc{ReviewBench} and injected the attack content directly into the main text of each paper.

\subsubsection{Attack Settings}
We consider the following three types of attacks.

\smallskip
\noindent \textbf{Disguised Misleading Information} injects plausible but entirely fabricated claims, such as fake state-of-the-art results, to test whether the system improperly relies on unverified statements. 

\smallskip
\noindent \textbf{Niche Terminology Induction} introduces undefined pseudo-technical terms to examine whether the model hallucinates explanations or incorrectly attributes novelty to them. 

\smallskip
\noindent \textbf{Scattered Key Information} fragments genuine key claims and disperses them across non-standard sections, testing the system's ability to aggregate evidence under challenging document organization.

\begin{table*}[t]
\centering
\caption{\textbf{Rubric-based evaluation under challenging adversarial attacks.} Higher scores indicate better performance. \textit{Contradict.} is a pitfall dimension scored in $\{-2,-1,0\}$, while all other dimensions are scored in $\{0,1,2\}$.}
\label{tab:appendix_adversarial_attacks}
\small
\setlength{\tabcolsep}{5pt}
\begin{tabular}{lccccccccc}
\toprule
Attack Type & Core & Res. & Comp. & EBC & Clr. & Cov. & Tone & Contradict. & Overall \\
\midrule
Disguised misleading information & 1.85 & 1.21 & 0.96 & 1.50 & 1.92 & 1.40 & 2.00 & -0.41 & 10.43 \\
Niche terminology induction & 1.78 & 1.39 & 0.91 & 1.63 & 1.95 & 1.41 & 2.00 & -0.26 & 10.81 \\
Scattered key information & 1.80 & 1.41 & 0.92 & 1.58 & 1.87 & 1.41 & 2.00 & -0.16 & 10.83 \\
Baseline (no attack) & \textbf{1.87} & 1.38 & \textbf{0.99} & \textbf{1.63} & 1.88 & \textbf{1.42} & \textbf{2.00} & \textbf{-0.17} & \textbf{11.00} \\
\bottomrule
\end{tabular}
\vspace{2mm}

\begin{minipage}{0.98\linewidth}
\footnotesize
\textbf{Abbreviations:}
Core = \textsc{Core Contribution Accuracy};
Res. = \textsc{Results Interpretation};
Comp. = \textsc{Comparative Analysis};
EBC = \textsc{Evidence-Based Critique};
Clr. = \textsc{Critique Clarity};
Cov. = \textsc{Completeness Coverage};
Tone = \textsc{Constructive Tone};
Contradict. = \textsc{False or Contradictory Claims}.
\end{minipage}
\end{table*}

\subsubsection{Result Interpretation}

As shown in Table~\ref{tab:appendix_adversarial_attacks}, \method maintains strong overall performance across all attack scenarios, achieving scores above 10.43 in every case, compared with 11.00 under the no-attack setting. Despite adversarial perturbations introducing moderate degradation on certain dimensions, particularly under the disguised misleading information setting, the overall drop remains limited. These results suggest that \method exhibits robust comprehension and strong resistance to subtle adversarial manipulations in paper content.

\subsection{Bias in Human Review Aggregation}

As noted in the reviews, ``human reviews can be noisy and do not always follow guidelines'', which motivates our use of aggregation to reduce such noise. While aggregating multiple reviews could introduce artifacts. In practice, it helps mitigate reviewer-specific biases and idiosyncrasies. Individual reviewers may emphasize different aspects of a paper or deviate from certain reviewing guidelines; by consolidating multiple reviews, our rubric construction process places greater weight on shared consensus and reduces the influence of outlier judgments. This yields more robust and stable reference reviews for constructing paper-specific rubrics. To further examine this issue, we perform an additional analysis under a single-review setting. Specifically, for each paper in \benchmark, we randomly select one of its human reviews as the sole reference review to instantiate the paper-specific rubric used for evaluating \method.

\begin{table*}[t]
\centering
\caption{\textbf{Rubric-based evaluation under the single human review setting.} For each paper, the rubric is instantiated from one randomly selected human review. Higher scores indicate better performance. \textit{Contradict.} is a pitfall dimension scored in $\{-2,-1,0\}$, while all other dimensions are scored in $\{0,1,2\}$.}
\label{tab:appendix_bias_aggregation}
\small
\setlength{\tabcolsep}{5pt}
\begin{tabular}{lccccccccc}
\toprule
Model & Core & Res. & Comp. & EBC & Clr. & Cov. & Tone & Contradict. & Overall \\
\midrule
Only Drafter (14B) & 1.6485 & 0.6260 & 0.4417 & 0.3616 & 1.7232 & 1.1812 & 1.9899 & -0.1896 & 7.7825 \\
ReviewGrounder & \textbf{1.8336} & \textbf{1.3771} & \textbf{0.7582} & \textbf{1.4962} & \textbf{1.8002} & \textbf{1.2201} & 1.9782 & \textbf{-0.1337} & \textbf{10.3298} \\
\bottomrule
\end{tabular}
\vspace{2mm}

\begin{minipage}{0.98\linewidth}
\footnotesize
\textbf{Abbreviations:}
Core = \textsc{Core Contribution Accuracy};
Res. = \textsc{Results Interpretation};
Comp. = \textsc{Comparative Analysis};
EBC = \textsc{Evidence-Based Critique};
Clr. = \textsc{Critique Clarity};
Cov. = \textsc{Completeness Coverage};
Tone = \textsc{Constructive Tone};
Contradict. = \textsc{False or Contradictory Claims}.
\end{minipage}
\end{table*}

As shown in Table~\ref{tab:appendix_bias_aggregation}, under the single human review setting, where each rubric is instantiated from only one randomly selected review, \method still substantially outperforms the 14B drafter baseline, achieving an overall score of 10.3298 compared to 7.7825. These gains indicate that our method remains robust even when rubric construction relies on only a single human review, further supporting the generality of the review-grounded refinement framework.

\subsection{Backbone Model Ablations}

We further analyze the impact of performance and computational overhead of diverse backbone model choices in \method by replacing specific modules with smaller backbone models and measuring the resulting performance degradation. 

In particular, we study two forms of efficiency-oriented ablations: (1) replacing the main drafter and grounding backbones with smaller models, and (2) replacing selected non-core grounding modules with 8B-scale models while keeping the rest of the system unchanged.

\subsubsection{Backbone Scaling Ablations}

We first replace the drafter \textsc{Phi-4-14B} with \textsc{Phi-4-7B} and replace the grounding model \textsc{GPT-OSS-120B} with \textsc{Qwen3-8B} to evaluate the performance impact of diverse backbone configurations.

\begin{table*}[t]
\centering
\caption{\textbf{Rubric-based evaluation under different backbone configurations.} Higher scores indicate better performance. \textit{Contradict.} is a pitfall dimension scored in $\{-2,-1,0\}$, while all other dimensions are scored in $\{0,1,2\}$ accordingly.}
\label{tab:ablation_backbone_scaling}
\small
\setlength{\tabcolsep}{4.2pt}
\resizebox{\textwidth}{!}{
\begin{tabular}{llcccccccccc}
\toprule
Model & Backbone & Core & Res. & Comp. & EBC & Clr. & Cov. & Tone & Contradict. & Overall & Est.\ VRAM \\
\midrule
Agent Review & GPT-4o & 1.0652 & 0.1321 & 0.1204 & 0.0348 & 1.2191 & 0.5585 & 1.8846 & -0.1472 & 4.8675 & - \\
AI Scientist & GPT-4o & 1.0052 & 0.0401 & 0.0244 & 0.0023 & 0.7021 & 0.1672 & 1.9477 & -0.2091 & 3.6800 & - \\
\method & P7+Q8 & 1.5443 & 0.5327 & 0.1672 & 0.0982 & 1.5327 & 0.6672 & 1.9899 & -0.3019 & 6.2302 & $\sim$30GB \\
\method & P14+Q8 & 1.5891 & 0.5844 & 0.2008 & 0.1502 & 1.5673 & 0.7136 & 1.9938 & -0.2111 & 6.5882 & $\sim$44GB \\
\method & P7+OSS & 1.8390 & 1.4400 & 0.8825 & 1.3700 & 1.8989 & 1.3219 & 1.9984 & -0.1579 & 10.5928 & $\sim$80GB \\
\method & P14+OSS & \textbf{1.8507} & 1.4075 & \textbf{0.9059} & \textbf{1.4831} & \textbf{1.9191} & \textbf{1.3289} & \textbf{1.9992} & \textbf{-0.1245} & \textbf{10.7699} & $\sim$108GB \\
\bottomrule
\end{tabular}
}
\vspace{2mm}

\begin{minipage}{0.98\linewidth}
\footnotesize
\textbf{Note:} ``P7/P14'' represents configuration ``Phi-4-7B or 14B'', ``Q8'' represents ``Qwen3-8B'', ``OSS'' reefer to ``GPT-OSS-120B''. \textsc{GPT-OSS-120B} uses native 4-bit MXFP4 quantization on its mixture-of-experts weights, which substantially reduces memory usage and makes deployment feasible on a single 80GB GPU in our staged setting.
\end{minipage}
\end{table*}

Table~\ref{tab:ablation_backbone_scaling} shows that \method remains competitive across a wide range of resource budgets. Even with the smallest configuration, \textsc{Phi-4-7B + Qwen3-8B} (approximately 30GB VRAM), the system achieves an overall score of 6.2302, substantially outperforming strong single-model baselines such as \textsc{Agent Review} and \textsc{AI Scientist}. Scaling the drafter from 7B to 14B under the 8B grounding setup yields a modest improvement (6.2302 $\rightarrow$ 6.5882), while preserving the large grounding model produces a much larger gain (e.g., 6.5882 $\rightarrow$ 10.7699), indicating that the grounding stage is the primary driver of final performance.

\subsubsection{Module-Level Ablations with Smaller Non-Core Models}

We then evaluate whether smaller models can replace selected non-core grounding modules without substantially harming end-to-end review quality. Individual modules are replaced in the grounding stage with \textsc{Qwen3-8B} and measure the resulting degradation on a random subset of 500 samples. 

\begin{table*}[t]
\centering
\caption{\textbf{Replacing selected non-core grounding modules with Qwen3-8B.} Results are reported on a random subset of 500 samples. Higher scores indicate better performance.}
\label{tab:ablation_module_small_models}
\small
\setlength{\tabcolsep}{5pt}
\resizebox{\textwidth}{!}{
\begin{tabular}{lccccccccc}
\toprule
Model & Core & Res. & Comp. & EBC & Clr. & Cov. & Tone & Contradict. & Overall \\
\midrule
14B + ablated insight miner & 1.8000 & 1.4060 & 0.8160 & 1.3659 & 1.8760 & 1.2560 & 1.9980 & -0.1340 & 10.3839 \\
14B + ablated paper summarizer & 1.8240 & 1.3620 & 0.8860 & 1.4336 & 1.9040 & 1.3040 & 2.0000 & -0.1360 & 10.5776 \\
14B + ablated result analyzer & 1.8280 & 1.1820 & 0.7600 & 1.4261 & 1.8740 & 1.3060 & 2.0000 & -0.1420 & 10.2341 \\
Original & \textbf{1.8300} & \textbf{1.4320} & 0.8460 & \textbf{1.4762} & \textbf{1.9240} & \textbf{1.3560} & \textbf{2.0000} & \textbf{-0.1240} & \textbf{10.7402} \\
\bottomrule
\end{tabular}
}
\end{table*}

As shown in Table~\ref{tab:ablation_module_small_models}, substituting the \emph{paper summarizer} leads to a minor drop in overall score (10.7402 $\rightarrow$ 10.5776), while ablating the \emph{insight miner} or \emph{result analyzer} causes larger but still moderate degradation. Among these, replacing the \emph{result analyzer} has the largest impact, especially on \textit{Res.} and \textit{Comp.}, suggesting that this module contributes more directly to accurate result interpretation and evidence integration.

To conclude, the results suggest a clear efficiency--performance trade-off. Smaller 7B/8B-scale models can potentially be used for the drafter or selected non-core grounding modules to substantially reduce computational overhead, while maintaining reasonable performance. However, retaining the large grounding backbone preserves most of the gains of \method, confirming that the grounding agents are the main source of improvement in review quality.

\section{Experimental Details}
\label{app:exp_details}
\subsection{Compared Baselines}
\label{app:baseline}
\begin{itemize}[leftmargin=1em]
% \item \textbf{Qwen3 Series}~\citep{yang2025qwen3}, developed by Alibaba, is a next-generation open-source LLM series available in dense (0.6B–32B) and Mixture-of-Experts variants (up to 235B). Qwen3 supports 119 languages and features a hybrid reasoning framework with a “Thinking Mode” for complex reasoning and a “Non-Thinking Mode” for efficient responses. The series excels in instruction following, reasoning, coding, long-context understanding, and tool use, achieving strong benchmark performance across NLP, coding, and math tasks. 
\item \textbf{Qwen3-32B}~\citep{yang2025qwen3}, a dense variant from Alibaba's Qwen3 Series, is used as a foundation model for review generation. It is optimized for instruction following, reasoning, and long-context understanding.

% \item \textbf{QwQ Series}~\citep{qwq32b}, developed by the Qwen Team at Alibaba, is a 32‑billion‑parameter reasoning model trained with scaled reinforcement learning to enhance analytical problem‑solving beyond conventional pretraining methods. The model achieves competitive performance with much larger reasoning models (e.g., DeepSeek‑R1) on benchmarks for mathematical reasoning, coding, and general problem solving. QwQ‑32B also integrates agent‑like capabilities that adapt reasoning based on environmental feedback.
\item \textbf{QwQ-32B}~\citep{qwq32b}, a reasoning model from Alibaba, is also used as a foundation model for review generation. It is trained to enhance analytical problem-solving and integrates agent-like reasoning capabilities.

% \item \textbf{GPT-4o Series}~\citep{hurst2024gpt}, produced by OpenAI, includes several model variants such as GPT-4o and GPT-4o-mini, with training leveraging extensive multimodal datasets encompassing text, vision, and audio modalities. The series achieves outstanding performance in complex reasoning tasks, creative generation, and multimodal understanding benchmarks with continuous refinements in alignment techniques and enhanced processing capabilities.
\item \textbf{GPT-4o}~\citep{hurst2024gpt}, an OpenAI model, demonstrates strong performance in complex reasoning, long-context comprehension, and creative generation. Its advanced reasoning and instruction-following capabilities enable it to produce coherent, detailed, and contextually informed outputs.

% \item \textbf{GPT‑4.1 Series}~\citep{openai_gpt41_2025}, released by OpenAI, comprises GPT‑4.1 and its smaller variants (GPT‑4.1 mini and GPT‑4.1 nano). These API‑only models outperform the previous GPT‑4o family across coding, instruction following, and long‑context understanding, supporting context windows up to 1 million tokens.
\item \textbf{GPT-4.1}~\citep{openai_gpt41_2025}, an OpenAI model, improves upon GPT-4o in coding, instruction following, and long-context understanding, supporting context windows up to 1 million tokens.

\item \textbf{AgentReview}~\citep{jin2024agentreview} models the academic peer-review process using LLM-driven agents that simulate reviewers, authors, and area chairs, enabling systematic analysis of reviewer bias, expertise, and decision dynamics without relying on real review data.

\item \textbf{AI Scientist}~\citep{lu2024ai} is an LLM-driven system that autonomously performs the entire scientific research process, including ideation, experiments, manuscript writing, and review, without human intervention. We adopt its review generation module as a baseline for evaluation.

\item \textbf{CycleReviewer}~\citep{weng2024cycleresearcher} is an LLM-based peer review simulator trained with iterative reinforcement learning to predict paper scores and generate review feedback. 

\item \textbf{DeepReviewer}~\citep{zhu2025deepreview} introduces a multi-stage LLM-driven paper review framework that emulates expert reviewers by combining structured analysis, literature retrieval, and evidence-based reasoning.

\end{itemize}

\subsection{Implementation Details}
\label{app:implementation_details}
We provide additional details on \method. In our main experiments, the \textit{Drafter} is instantiated with Phi-4-14B~\citep{abdin2024phi}, while the other modules, including the \textit{Literature Searcher}, \textit{Insight Miner}, \textit{Result Analyzer}, and \textit{Aggregator}, are instantiated with GPT-OSS-120B. For paper reranking in $\mathcal{S}$, we adopt OpenScholar-Reranker~\citep{asai2024openscholar}. Only the \textit{Drafter} is trainable. The model is trained on a portion of the DeepReview-13K dataset using 8 NVIDIA A100 80GB GPUs with Model-Swift~\citep{zhao2025swift}, DeepSpeed, and ZeRO-3 optimization~\citep{rajbhandari2020zero}. Training is performed for three epochs with a batch size of 16 and a learning rate of $5\times10^{-6}$. During \method review generation, we set the temperature to $0.4$ and the maximum input and output lengths to 100K and 16,384 tokens, respectively, to ensure full text coverage.

\section{More Discussion about Experiment Results}
\subsection{Main Result Analysis}
\label{app:main_results_analysis}
Our main results are presented in Table~\ref{tab:main_rubric}. Overall, \method consistently outperforms all baseline models across all rubric dimensions. These comprehensive results yield several key insights:

\smallskip
\noindent \textbf{Foundation models are insufficient for high-quality review generation.} While strong foundation models such as GPT-4.1, GPT-4o achieve near-ceiling performance on surface-level criteria (e.g., \textsc{Constructive Tone} and  \textsc{Critique Clarity}), they consistently underperform on core analytical and contextual-grounding dimensions, including \textsc{Evidence-Based Critique}, \textsc{Comparative Analysis}, and \textsc{Results Interpretation}. In contrast, our proposed \method substantially improves performance on these dimensions, achieving 1.4831 on \textsc{Evidence-Based Critique} (vs. 0.0024 for GPT-4o), 0.9059 on \textsc{Comparative Analysis} (vs. 0.3406 for GPT-4.1), and 1.4075 on \textsc{Results Interpretation} (vs. 0.1037 for GPT-4o). This imbalance results in substantially lower overall scores compared to \method, indicating that single-pass generation fails to meet the requirements of rigorous academic peer review.

\smallskip
\noindent \textbf{Agentic frameworks and fine-tuned models yield improvements in specific dimensions but limited generalization.} Agentic reviewer systems, such as AgentReview with GPT-4o, compared with its backbone, improve \textsc{Completeness Coverage} (0.5900 vs. 0.3318) and \textsc{Critique Clarity} (1.3400 vs. 1.0499), while fine-tuned models, including DeepReviewer based on Phi-4-14B, achieve moderate gains on \textsc{Evidence-Based Critique} (0.3532), \textsc{Comparative Analysis} (0.4977), and \textsc{Results Interpretation} (0.6532). Nonetheless, both agentic and fine-tuned approaches still underperform on core analytical and contextual-grounding dimensions, leaving substantial gaps relative to \method in producing high-quality, balanced reviews.

\smallskip
\noindent \textbf{\method delivers high-quality, evidence-grounded, and substantive reviews with critical insights.} ReviewGrounder establishes a new state-of-the-art in automatic peer review by achieving an overall score of 10.7699, with particularly strong performance on key rubric dimensions: 1.8507 on \textsc{CORE CONTRIBUTION ACCURACY}, 1.4831 on \textsc{Evidence-Based Critique}, 1.4075 on \textsc{Results Interpretation}, and 1.9992 on \textsc{CONSTRUCTIVE TONE}. This underscores that the core advantage of ReviewGrounder comes from its tool-integrated, rubric-guided framework.

\subsection{Detailed Analysis of Ablation Results}
\label{app:ablation_study}

% \begin{figure}[t]
%     \centering
%     \includegraphics[width=\linewidth]{figs/ablation-app.png}
%     \caption{\textbf{Fine-grained dimensional breakdown of the ablation study.} We report the detailed score distribution across 8 rubric dimensions for all configurations, covering component ablation (removing Analyzer, Miner, Searcher), Drafter backbone scaling, searcher reranking settings ($N \in \{5, 10, 20\}$), and retriever selection. The numbers inside the bars represent the weighted contribution of each dimension, and the box on the right displays the final Overall Score.}
%     \label{fig:ablation_detailed}
%     \vspace{-1em}
% \end{figure}

\begin{figure}[h]
    \centering
    \includegraphics[width=\linewidth]{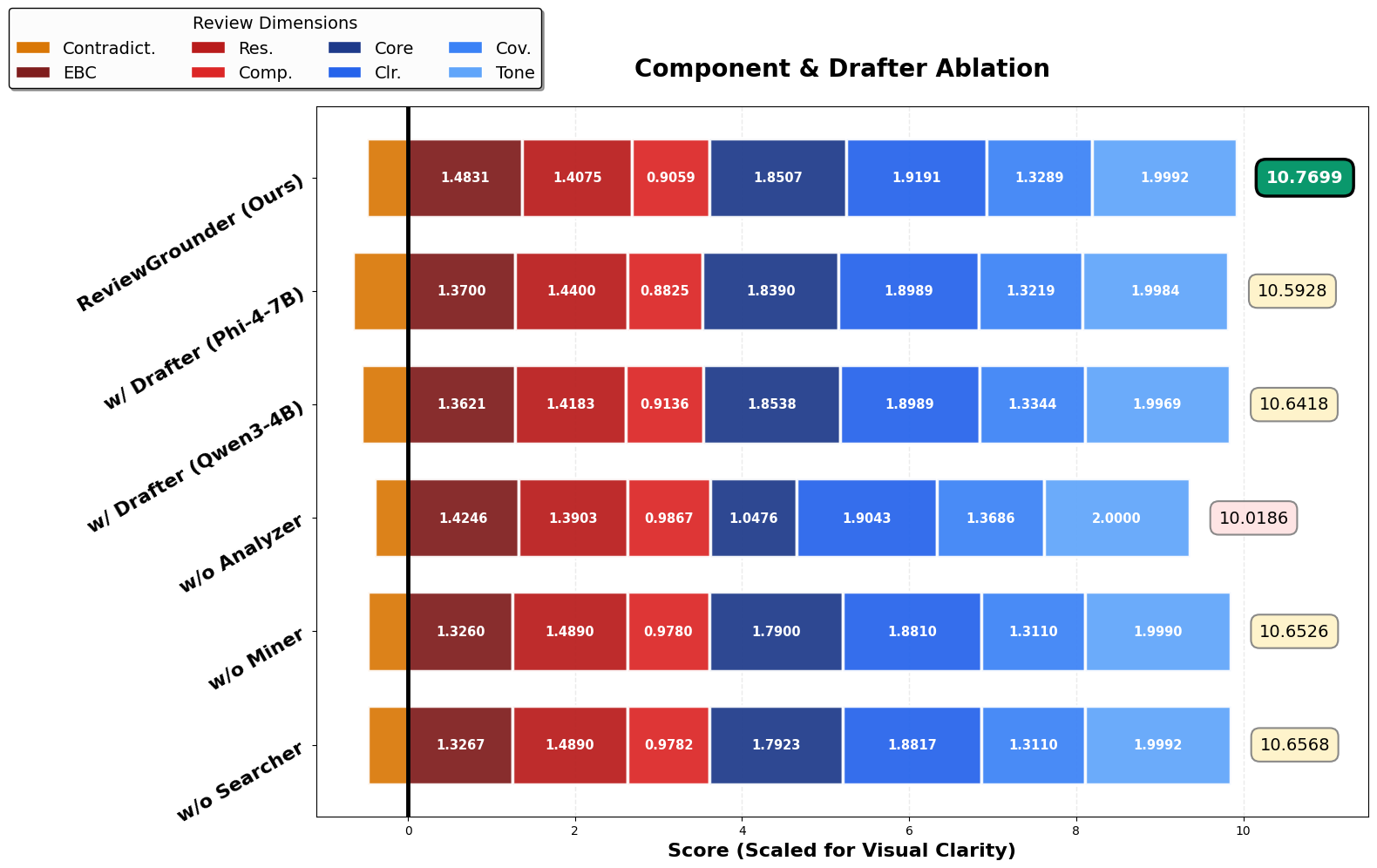}
    \caption{\textbf{Component \& Drafter Ablation Study.} Fine-grained score breakdown across 8 rubric dimensions, analyzing the impact of removing system components (Result Analyzer, Insight Miner, Literature Searcher) and scaling the Drafter backbone.
    }
    \label{app:fig:ablation_component}
    \vspace{-0.5em}
\end{figure}

\begin{figure}[h]
    \centering
    \includegraphics[width=\linewidth]{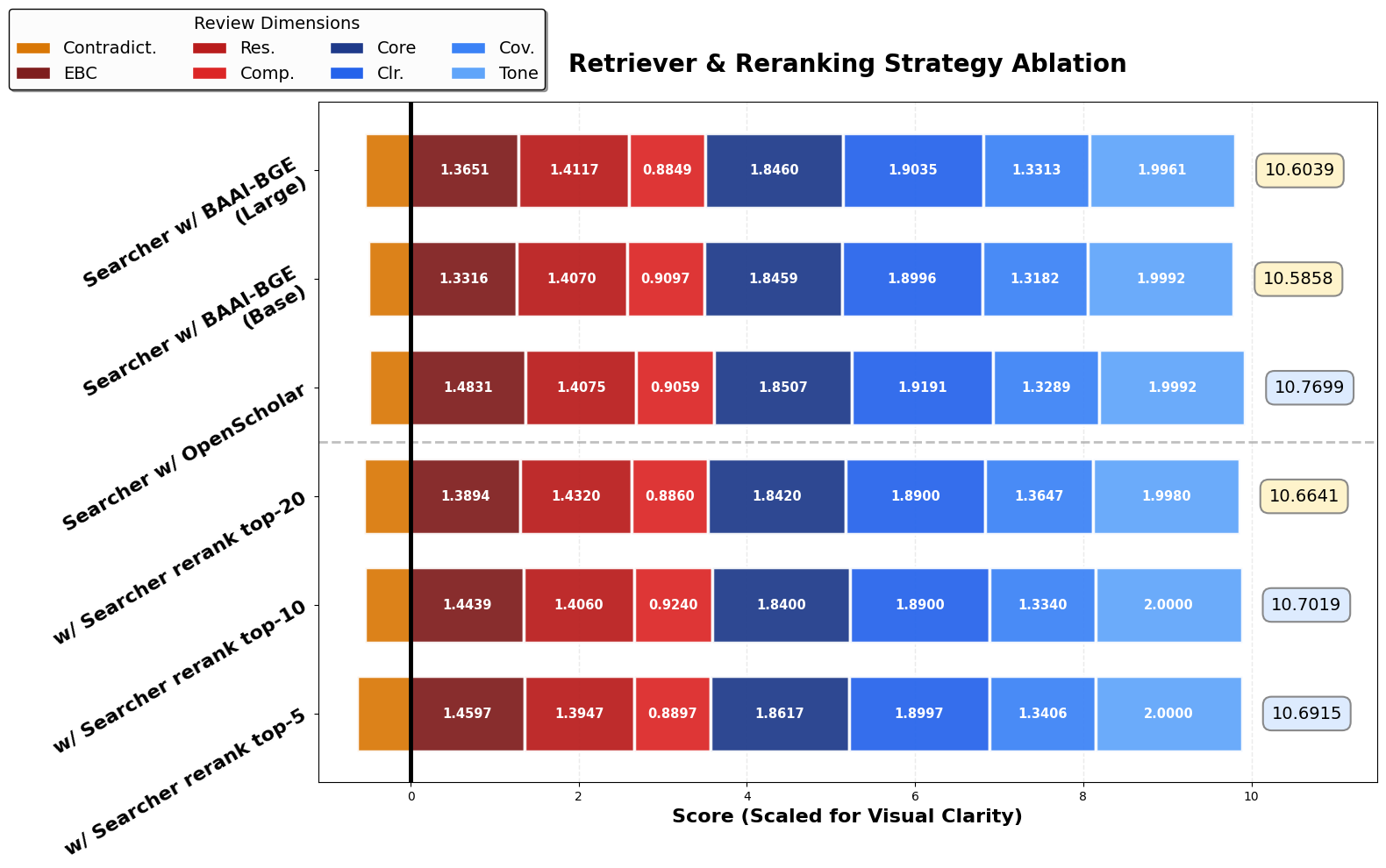}
    \caption{\textbf{Retriever \& Reranking Strategy Ablation.} Detailed performance comparison across retriever selections (OpenScholar vs. BAAI-BGE) and different reranking settings ($N \in \{5, 10, 20\}$).
    }
    \label{app:fig:ablation_retriever}
    \vspace{-1em}
\end{figure}

To further understand where the performance gains originate, Figure~\ref{app:fig:ablation_component} and Figure~\ref{app:fig:ablation_retriever} visualize the score breakdown across the eight evaluation dimensions.

\smallskip
\noindent \textbf{The Critical Role of the Result Analyzer.} The most striking observation is the degradation caused by removing the Result Analyzer (w/o Analyzer), which results in the lowest overall score of 10.086. Visually, this configuration suffers from a significant contraction in the \textsc{Core Contribution Accuracy} (1.0476) and \textsc{Results Interpretation} (1.3903) dimensions. This empirically validates that without a specialized agent to verify experimental data against the paper's tables and figures, the system fails to produce substantiated critiques, leading to superficial reviews.

\smallskip
\noindent \textbf{Resilience of Smaller Drafters.} Comparing the Drafter backbones, while the Phi-4-14B achieves the state-of-the-art score (10.7699), the smaller Qwen3-4B backbone (10.6418) surprisingly outperforms the mid-sized Phi-4-7B (10.5928). This suggests that our multi-agent grounding mechanism effectively compensates for the reasoning limitations of smaller models (like the 4B parameter model), boosting their ability to perform comparative analysis and verify claims even with a weaker base generator.

\subsection{Hyperparameter Study}

\noindent \textbf{Optimization of Information Retrieval. }In the Literature Searcher $\mathcal{S}$, the breakdown shows that the performance gap between OpenScholar and BAAI-BGE is distributed across multiple dimensions, particularly \textit{Evidence Critique}. This confirms that domain-specific retrieval is not just about finding papers, but about finding the right context to position the submission correctly. Furthermore, the reranking ablation ($N=10$ vs. $N=5/20$) illustrates a trade-off: insufficient context ($N=5$) hampers the \textit{Comparative Analysis} dimension, while excessive context ($N=20$) introduces noise that slightly degrades \textit{Critique Clarity}.

\clearpage
% \onecolumn

\section{Rubrics in \benchmark}
\label{app:rubrics}

\subsection{Meta Rubrics}
\label{app:meta_rubrics}
\begin{promptbox}[title=Meta Rubrics]{toolcardborder}

\textbf{Purpose:}  
These rubrics define the evaluation criteria used to assess the \emph{quality, correctness, and usefulness} of a peer review. Each dimension focuses on a specific aspect of review quality and guides both automated refinement agents and human supervisors.

\vspace{6pt}

\noindent\textbf{Evaluation Dimensions:}

\begin{itemize}[left=0pt, itemsep=6pt, parsep=0pt]

  \item \textbf{Core Contribution Accuracy} \hfill \\
  Assesses whether the review accurately captures the paper’s main contributions and central methodological innovations, especially in the \textit{Summary} and \textit{Strengths}, without misinterpretation.

  \item \textbf{Results Interpretation} \hfill \\
  Evaluates whether the review correctly interprets empirical results, including tables, figures, metrics, and statistical comparisons, rather than overstating or misreading findings.

  \item \textbf{Comparative Analysis} \hfill \\
  Checks whether the review appropriately discusses comparisons with baselines and related work that are actually presented in the paper, and avoids unsupported claims about missing comparisons.

  \item \textbf{Evidence-Based Critique} \hfill \\
  Examines whether criticisms and weaknesses are grounded in verifiable paper content, such as specific sections, equations, algorithms, tables, or figures, rather than vague impressions.

  \item \textbf{Critique Clarity} \hfill \\
  Determines whether weaknesses and questions are stated clearly and concretely enough for authors to understand what needs improvement and how to address it.

  \item \textbf{Completeness Coverage} \hfill \\
  Assesses whether the review covers all major aspects of the paper, including methodology, theoretical formulation, experimental evaluation, and related-work positioning.

  \item \textbf{Constructive Tone} \hfill \\
  Evaluates whether the review maintains a professional, constructive, and improvement-oriented tone, rather than being dismissive or discouraging.

  \item \textbf{False or Contradictory Claims} \hfill \\
  Penalizes reviews that mention experiments or content absent from the paper, incorrectly claim something is missing when it exists, or contradict the paper’s stated results, conclusions, or explicit design choices.

\end{itemize}

\vspace{6pt}

\noindent\textbf{Note:}  
The \emph{False or Contradictory Claims} dimension represents a critical failure mode that should be strictly avoided.
\end{promptbox}

\subsection{Paper-specific Rubrics}
We provide paper-specific rubrics below for \textit{SqueezeLLM: Dense-and-Sparse Quantization}~\citep{kim2023squeezellm} as a case study.
 
\begin{promptbox}[title=Paper-Specific Review Rubrics:\\ Paper \textit{SqueezeLLM}]{toolcardborder}

\textbf{Purpose:}  
These rubrics are tailored to the paper \textit{SqueezeLLM: Dense and Sparse Quantization}.  
It specifies how a review should be evaluated \emph{with respect to this paper’s concrete claims, methods, and results}.  
This rubric is for \textbf{display and guidance only}, not for automated scoring.

\vspace{6pt}

\noindent\textbf{Paper-Aware Evaluation Dimensions:}

\begin{itemize}[left=0pt, itemsep=6pt, parsep=0pt]

  \item \textbf{Core Contribution Accuracy}. 
  Checks whether the review correctly captures the paper’s two primary technical contributions:
  (i) sensitivity-based non-uniform quantization using Fisher-information-weighted k-means, and  
  (ii) dense-and-sparse decomposition for handling outliers and sensitive weights.  
  The review should also accurately reflect the paper’s central claim that \emph{memory bandwidth, not compute}, is the main bottleneck for single-batch LLM inference.

  \item \textbf{Results Interpretation}. 
  Evaluates whether the review correctly interprets the reported empirical results, including:
  perplexity comparisons at 3-bit and 4-bit precision, latency speedups on A6000 GPUs, and accuracy results on MMLU and Vicuna benchmarks.  
  Numeric claims should align with the tables and figures presented in the paper.

  \item \textbf{Comparative Analysis}. 
  Assesses whether the review properly discusses comparisons with prior PTQ methods explicitly evaluated in the paper (e.g., RTN, GPTQ, AWQ, SpQR),
  and whether it reflects the authors’ claims regarding superiority in low-bit performance and inference efficiency.

  \item \textbf{Evidence-Based Critique}.
  Requires that critiques of the paper (e.g., computational cost of Fisher-information estimation, kernel efficiency, missing ablations)  
  are grounded in concrete references such as specific sections, equations, tables, or figures (e.g., Eq.~2--4, Fig.~2, Table~1).

  \item \textbf{Critique Clarity}.
  Evaluates whether weaknesses and questions are stated with sufficient specificity (e.g., identifying which model sizes, tables, or steps are affected),  
  enabling the authors to clearly understand what needs clarification or additional evidence.

  \item \textbf{Completeness Coverage}.
  Determines whether the review covers all major components of the paper, including:
  memory-wall motivation, quantization methodology, theoretical formulation, experimental evaluation, kernel implementation, and related-work discussion.

  \item \textbf{Constructive Tone}.
  Assesses whether the review maintains a professional and constructive tone, balancing acknowledgment of strengths with critical feedback.

  \item \textbf{False or Contradictory Claims}.
  Penalizes reviews that:
  misstate the paper’s methods or results, claim missing experiments that are actually present,  
  or contradict the paper’s explicitly stated design choices or reported findings.

\end{itemize}

\vspace{6pt}

\noindent\textbf{Note:}  
This paper-specific rubric complements the general meta-rubric by anchoring evaluation criteria directly to the claims, methods, and evidence presented in \textit{SqueezeLLM}.

\end{promptbox}

\section{Instruction Templates}
\label{app:prompt}
\subsection{\benchmark}
\label{app:prompt_reviewbench}

\subsubsection{Evaluator}
\begin{promptbox}[title=Scoring Rules]{toolcardborder}
\label{box:scoring_rules}
\textbf{Positive dimension ($0/1/2$).}
Let $\hat{R}$ denote the paper-specific rule and let $\mathcal{K}(\hat{R})$ be its checklist of key points.
\begin{itemize}
  \item \textbf{Score 0 (not satisfied):} The review addresses none of the key points, \emph{or} it makes a \emph{material} incorrect claim relevant to this dimension.
  \item \textbf{Score 1 (partially satisfied):} The review correctly addresses at least half of the key points and contains no material errors.
  \item \textbf{Score 2 (fully satisfied):} The review correctly addresses all key points (or all but a minor omission that does not affect the criterion) and contains no material errors.
\end{itemize}

\textbf{Negative pitfall dimension ($-2/-1/0$).}
\begin{itemize}
  \item \textbf{Score 0 (none):} The review exhibits none of the pitfall key points.
  \item \textbf{Score $-1$ (mild):} The review exhibits at least one pitfall key point.
  \item \textbf{Score $-2$ (severe):} The review exhibits multiple pitfall key points, \emph{or} a single severe instance (e.g., a clear hallucination of nonexistent content or a direct contradiction of stated results/design choices).
\end{itemize}
\end{promptbox}

\subsection{\method}
\label{app:prompt_reviewgrounder}

\subsubsection{Drafter}
\begin{promptbox}[title=Instruction for {Drafter}]{toolcardborder}
\textbf{Task:} Provide fair, thorough, and constructive evaluations of research papers, highlighting summary, strengths, weaknesses, and questions.

\vspace{6pt}

\noindent\textbf{Role / Prompt:}
You are an expert academic reviewer tasked with providing a thorough and balanced evaluation of research papers.

\vspace{6pt}

\noindent\textbf{Inputs:}
\begin{itemize}[left=0pt, label={}, itemsep=0pt, parsep=0pt]
  \item Paper Information: \smalltt{\{context\}}
\end{itemize}

% \vspace{6pt}

% \noindent\textbf{What to do:}
% \begin{enumerate}[left=0pt, itemsep=0pt, parsep=0pt]
%   \item Identify what the related work is about and its main contributions.
%   \item Summarize the main methods used in the related work.
%   \item Summarize the key results or findings reported in the related work.
%   \item Explain the relationship between the related work and the reference paper, focusing on:
%   \begin{itemize}[left=0pt, label={}, itemsep=0pt, parsep=0pt]
%     \item shared ideas or problem settings,
%     \item differences in methods or assumptions,
%     \item complementary or diverging claims.
%   \end{itemize}
% \end{enumerate}

% \vspace{6pt}

% \noindent\textbf{Guidelines:}
% \begin{itemize}[left=0pt, label={}, itemsep=0pt, parsep=0pt]
%   \item Focus on the \textbf{relationship} between the two papers rather than standalone details.
%   \item Be concise and informative; avoid unnecessary background.
%   \item Do NOT add external knowledge beyond the provided papers.
% \end{itemize}

% \vspace{6pt}

% \noindent\textbf{Output JSON only:}
% \begin{codebox}
% {
%   "summary": "Your concise summary here in 2-3 sentences.",
%   "main_methods": "The main methods of the related work.",
%   "key_findings": "The key findings of the related work.",
%   "relation": "The relation between the related work and the paper you are reviewing, such as shared ideas, similar problems, or diverging approaches."
% }
% \end{codebox}
\end{promptbox}

\subsubsection{Literature Searcher}
\begin{promptbox}[title=Instruction for {Keyword Generation}]{toolcardborder}
\textbf{Task:} Generate concise search queries (keywords) to retrieve relevant related-work papers based on the given paper information.

\vspace{6pt}

\noindent\textbf{Role / Prompt:}
You are an experienced research assistant helping to find related work for a paper by identifying core technical concepts, methods, and key techniques.

\vspace{6pt}

\noindent\textbf{Input:}
\begin{itemize}[left=0pt, label={}, itemsep=0pt, parsep=0pt]
  \item Paper information: \smalltt{\{context\}}
\end{itemize}

\vspace{6pt}

\noindent\textbf{What to do:}
\begin{enumerate}[left=0pt, itemsep=0pt, parsep=0pt]
  \item Identify the paper’s core technical concepts, methods, and key techniques.
  \item Generate short, simple search queries suitable for academic literature search.
  \item Prefer general, reusable phrases rather than overly specific titles.
\end{enumerate}

\vspace{6pt}

\noindent\textbf{Guidelines:}
\begin{itemize}[left=0pt, label={}, itemsep=0pt, parsep=0pt]
  \item Generate \textbf{3--5} keywords only.
  \item Each keyword should be short and concise.
  \item Keywords should be suitable as standalone search queries.
  \item Do NOT include explanations or commentary.
\end{itemize}

\vspace{6pt}

\noindent\textbf{Output JSON only:}
\begin{codebox}
{
  "keywords": ["keyword1", "keyword2", "keyword3", "keyword4", "keyword5"]
}
\end{codebox}
\end{promptbox}

\begin{promptbox}[title=Instruction for \\{Related-Work Summarization}]{toolcardborder}
\textbf{Task:} Generate a concise, structured summary of a related paper, focusing on its main methods, key findings, and its relationship to a given reference paper.

\vspace{6pt}

\noindent\textbf{Role / Prompt:}
You are a senior research assistant proficient at identifying the main contributions, key findings of papers, and the relationships between different works.

\vspace{6pt}

\noindent\textbf{Inputs:}
\begin{itemize}[left=0pt, label={}, itemsep=0pt, parsep=0pt]
  \item Reference paper: \smalltt{\{reference\_paper\}}
  \item Related paper: \smalltt{\{related\_paper\}}
\end{itemize}

\vspace{6pt}

\noindent\textbf{What to do:}
\begin{enumerate}[left=0pt, itemsep=0pt, parsep=0pt]
  \item Identify what the related work is about and its main contributions.
  \item Summarize the main methods used in the related work.
  \item Summarize the key results or findings reported in the related work.
  \item Explain the relationship between the related work and the reference paper, focusing on:
  \begin{itemize}[left=0pt, label={}, itemsep=0pt, parsep=0pt]
    \item shared ideas or problem settings,
    \item differences in methods or assumptions,
    \item complementary or diverging claims.
  \end{itemize}
\end{enumerate}

\vspace{6pt}

\noindent\textbf{Guidelines:}
\begin{itemize}[left=0pt, label={}, itemsep=0pt, parsep=0pt]
  \item Focus on the \textbf{relationship} between the two papers rather than standalone details.
  \item Be concise and informative; avoid unnecessary background.
  \item Do NOT add external knowledge beyond the provided papers.
\end{itemize}

\vspace{6pt}

\noindent\textbf{Output JSON only:}
\begin{codebox}
{
  "summary": "Your concise summary here in 2-3 sentences.",
  "main_methods": "The main methods of the related work.",
  "key_findings": "The key findings of the related work.",
  "relation": "The relation between the related work and the paper you are reviewing, such as shared ideas, similar problems, or diverging approaches."
}
\end{codebox}
\end{promptbox}

\subsubsection{Insight Miner}

\begin{promptbox}[title=Instruction for {Insight Miner}]{toolcardborder}
\textbf{Task:} Refine the \textbf{method and contribution} parts of a candidate review using the paper text as the sole source of truth, and provide paper-grounded rewrite suggestions with concrete evidence.

\vspace{6pt}

\noindent\textbf{Role / Prompt:}
You are an expert research assistant. Your task is to help refine the method/contribution parts of a candidate review, using the paper content as the source of truth.

\vspace{6pt}

\noindent\textbf{SCOPE (strict):}
\begin{itemize}[left=0pt, label={}, itemsep=0pt, parsep=0pt]
  \item ONLY cover: core contributions, technical approach, model/algorithm design, mathematical formulation, assumptions, optimization/training, implementation details, and method limitations.
  \item Novelty: ONLY assess novelty claims \textbf{as presented in the paper itself} (no external knowledge, no web search).
  \item Do NOT comment on experimental results, benchmarks, or score/decision fields.
  \item Do NOT do external related-work positioning.
\end{itemize}

\vspace{6pt}

\noindent\textbf{Inputs:}
\begin{itemize}[left=0pt, label={}, itemsep=0pt, parsep=0pt]
  \item Paper content: \smalltt{\{content\}}
  \item Candidate review: \smalltt{\{candidate\_review\}}
\end{itemize}

\vspace{6pt}

\noindent\textbf{What to do:}
\begin{enumerate}[left=0pt, itemsep=0pt, parsep=0pt]
  \item Extract the paper’s core contributions and method details (paper-grounded).
  \item Check the candidate review’s method/contribution claims and identify:
  \begin{itemize}[left=0pt, label={}, itemsep=0pt, parsep=0pt]
    \item incorrect / hallucinated / contradicted claims,
    \item missing key technical points,
    \item vague or generic statements that should be made specific.
  \end{itemize}
  \item Provide short rewrite suggestions \textbf{WITH evidence anchors} (Section / Equation / Algorithm / Figure / snippet if available).
\end{enumerate}

\vspace{6pt}

\noindent\textbf{Rules:}
\begin{itemize}[left=0pt, label={}, itemsep=0pt, parsep=0pt]
  \item If you cannot find support in the paper text, set \smalltt{evidence} to \smalltt{"not\_found\_in\_text"}; do NOT assert the paper is missing it.
  \item Keep each list short ($\le$ 5 items). Prefer the most important contributions/components/issues.
  \item Return JSON only. No extra text.
\end{itemize}

\vspace{6pt}

\noindent\textbf{Output JSON only:}
\begin{codebox}
{
  "facts": {
    "core_contributions": [
      {"claim": "...", "evidence": "..."}
    ],
    "method_summary": [
      {"point": "key component / step / design choice", "evidence": "..."}
    ],
    "assumptions_and_scope": [
      {"item": "...", "evidence": "..."}
    ],
    "novelty_claims_in_paper": [
      {"claim": "as stated by the authors", "evidence": "..."}
    ]
  },
  "review_issues": {
    "incorrect_or_hallucinated": [
      {"review_claim": "...", "why_wrong": "...", "evidence": "..."}
    ],
    "missing_key_points": [
      {"what_missing": "...", "why_important": "...", "evidence": "..."}
    ],
    "needs_specificity": [
      {"review_text": "...", "how_to_fix": "name the component/assumption/equation/step", "evidence": "..."}
    ]
  },
  "rewrite_suggestions": [
    {
      "apply_to": "Summary|Strengths|Weaknesses|Questions (method-related only)",
      "target": "Core Contribution Accuracy|Evidence-Based Critique|Critique Clarity",
      "suggested_text": "1-2 sentences",
      "evidence": "..."
    }
  ]
}
\end{codebox}
\end{promptbox}

\subsubsection{Result Analyzer}
\begin{promptbox}[title=Instruction for {Result Analyzer}]{toolcardborder}
\textbf{Task:} Pinpoint issues in the experiment/evaluation parts of a candidate review using the paper text as the source of truth, and provide paper-grounded rewrite suggestions with concrete evidence.

\vspace{6pt}

\noindent\textbf{Role / Prompt:}
You are an expert research assistant. Your task is to help refine the experiment/evaluation parts of a candidate review, using the paper content as the source of truth.

\vspace{6pt}

\noindent\textbf{SCOPE (strict):}
\begin{itemize}[left=0pt, label={}, itemsep=0pt, parsep=0pt]
  \item ONLY cover experimental evaluation: datasets, baselines, metrics, tables/figures, quantitative results, statistical evidence, ablations.
  \item Do NOT comment on novelty, related-work positioning, writing/presentation quality, or overall recommendation. Other agents will handle those.
\end{itemize}

\vspace{6pt}

\noindent\textbf{Inputs:}
\begin{itemize}[left=0pt, label={}, itemsep=0pt, parsep=0pt]
  \item Paper content: \smalltt{\{content\}}
  \item Candidate review: \smalltt{\{candidate\_review\}}
\end{itemize}

\vspace{6pt}

\noindent\textbf{What to do:}
\begin{enumerate}[left=0pt, itemsep=0pt, parsep=0pt]
  \item Extract key experimental facts from the paper.
  \item Check experiment-related claims in the candidate review and identify:
  \begin{itemize}[left=0pt, label={}, itemsep=0pt, parsep=0pt]
    \item incorrect/hallucinated/contradicted claims,
    \item missing key experimental points,
    \item vague statements that should be made specific.
  \end{itemize}
  \item Provide short rewrite suggestions \textbf{WITH evidence anchors} (Table/Figure/Section/snippet if available).
\end{enumerate}

\vspace{6pt}

\noindent\textbf{Rules:}
\begin{itemize}[left=0pt, label={}, itemsep=0pt, parsep=0pt]
  \item If you cannot find support in the paper text, set \smalltt{evidence} to \smalltt{"not\_found\_in\_text"}; do NOT assert the paper is missing it.
  \item Keep each list short ($\le$ 5 items). Prefer the most important issues/results.
  \item Return JSON only. No extra text.
\end{itemize}

\vspace{6pt}

\noindent\textbf{Output JSON only:}
\begin{codebox}
{ 
  "facts": { 
    "datasets": ["..."], 
    "metrics": ["..."],
    "baselines": ["..."],
    "key_results": [
      {"claim": "...", "evidence": "..."}
    ]
  },
  "review_issues": {
    "incorrect_or_hallucinated": [
      {"review_claim": "...", "why_wrong": "...", "evidence": "..."}
    ],
    "missing_key_points": [
      {"what_missing": "...", "why_important": "...", "evidence": "..."}
    ],
    "needs_specificity": [
      {"review_text": "...", "how_to_fix": "...", "evidence": "..."}
    ]
  },
  "rewrite_suggestions": [
    {
      "apply_to": "Summary|Strengths|Weaknesses|Questions (experiment-related only)",
      "target": "Results Interpretation|Evidence-Based Critique",
      "suggested_text": "1-2 sentences",
      "evidence": "..."
    }
  ]
}
\end{codebox}
\end{promptbox}

\subsubsection{Aggregator}
\begin{promptbox}[title=Instruction for {Aggregator}]{toolcardborder}
\textbf{Task:} Refine an existing peer review to improve factual grounding, coverage, and usefulness while preserving the draft’s structure and intent. Treat the paper text as the source of truth.

\vspace{6pt}

\noindent\textbf{Role / Prompt:}
You are a senior researcher refining an existing peer review. Your job is to improve factual grounding, coverage, and usefulness while preserving the draft’s structure and intent. Treat the paper text as the source of truth.

\vspace{6pt}

\noindent\textbf{You will be given:}
\begin{enumerate}[left=0pt, itemsep=0pt, parsep=0pt]
    \item Paper text (plain text converted from PDF)
    \item Draft review (structured)
    \item Method/Contribution audit report (from Paper Insight Miner; paper-grounded)
    \item Experiments/Results audit report (from Paper Results Analyzer; paper-grounded)
    \item Related-work summaries (each item is a JSON summary of one retrieved paper, written relative to the target paper)
\end{enumerate}

\vspace{6pt}

\noindent\textbf{Primary objectives (what to improve):}
Refine the review to satisfy these content-quality dimensions:
\begin{enumerate}[left=0pt, itemsep=0pt, parsep=0pt]
    \item Core Contribution Accuracy
    \item Results Interpretation
    \item Comparative Analysis / Positioning
    \item Evidence-Based Critique
    \item Critique Clarity
    \item Completeness Coverage
    \item Constructive Tone
    \item Avoid False or Contradictory Claims (critical)
\end{enumerate}

\vspace{6pt}

\noindent\textbf{Hard constraints (must follow):}
\begin{enumerate}[left=0pt, itemsep=2pt, parsep=0pt]
    \item \textbf{Paper-grounded correctness is mandatory:}
    \begin{itemize}[left=0pt, label={}, itemsep=0pt, parsep=0pt]
        \item If the audit reports mark a draft claim as incorrect/hallucinated/contradicted, you \textbf{MUST} fix or remove it.
        \item Do \textbf{NOT} introduce new factual claims about the paper unless you can anchor them to the paper text or the audit reports’ evidence.
    \end{itemize}

    \item \textbf{Evidence anchoring rule:}
    \begin{itemize}[left=0pt, label={}, itemsep=0pt, parsep=0pt]
        \item Every major critique (esp. in Weaknesses/Questions) must include a verifiable anchor:
        section name, table/figure identifier, equation/algorithm reference, dataset/metric name, or a short quote snippet ($\le$ 20 words).
        \item If you cannot find support, convert the statement into a question or a suggestion for clarification (do not assert absence).
    \end{itemize}

    \item \textbf{Related-work usage rule (anti-leak / anti-overclaim):}
    \begin{itemize}[left=0pt, label={}, itemsep=0pt, parsep=0pt]
        \item Retrieved related-work summaries are \textbf{NOT} guaranteed to be cited by the submission.
        \item Never claim ``the paper compares to/cites X'' unless the paper text actually contains X.
        \item When using retrieved works, attribute them as external context:
        ``The related-work search suggests ...; it would help to clarify/compare ...''
        \item Use related work to: (i) sharpen positioning, (ii) propose missing baselines/comparisons, (iii) raise targeted questions.
    \end{itemize}

    \item \textbf{Minimal-change policy:}
    \begin{itemize}[left=0pt, label={}, itemsep=0pt, parsep=0pt]
        \item Keep the original structure and as much of the draft wording as possible.
        \item Do \textbf{NOT} shorten aggressively; do \textbf{NOT} rewrite into a totally new review.
        \item Prefer targeted edits, insertions, and corrections.
    \end{itemize}

    \item \textbf{Numeric fields policy (IMPORTANT):}
    \begin{itemize}[left=0pt, label={}, itemsep=0pt, parsep=0pt]
        \item Default: keep \textbf{ALL} numeric fields and the decision unchanged.
        \item Change numeric fields \textbf{ONLY} if the refined textual assessment would otherwise be clearly inconsistent, or if a major factual correction materially changes the evaluation.
        \item If you change any numeric field: change the minimum number of fields, and keep changes small unless necessary.
    \end{itemize}
\end{enumerate}

\vspace{6pt}

\noindent\textbf{How to use the tool reports (operational):}
\begin{enumerate}[left=0pt, itemsep=2pt, parsep=0pt]
    \item \textbf{Apply Paper Insight Miner (method/contribution):}
    \begin{itemize}[left=0pt, label={}, itemsep=0pt, parsep=0pt]
        \item Use \smalltt{review\_issues.incorrect\_or\_hallucinated} to remove/correct wrong claims in Summary/Strengths/Weaknesses.
        \item Use \smalltt{missing\_key\_points} and \smalltt{needs\_specificity} to improve technical specificity.
        \item Incorporate \smalltt{rewrite\_suggestions} where appropriate (method-related only).
    \end{itemize}

    \item \textbf{Apply Paper Results Analyzer (experiments/results):}
    \begin{itemize}[left=0pt, label={}, itemsep=0pt, parsep=0pt]
        \item Correct any wrong result interpretation.
        \item Add missing datasets/baselines/metrics/key results if they are important and supported.
        \item Convert vague experiment critiques into concrete, testable suggestions with anchors.
        \item Incorporate \smalltt{rewrite\_suggestions} where appropriate (experiment-related only).
    \end{itemize}

    \item \textbf{Use Related-work summaries:}
    \begin{itemize}[left=0pt, label={}, itemsep=0pt, parsep=0pt]
        \item Use each item’s \smalltt{relation} to craft 1--3 concrete positioning points:
        what is similar/different, what comparisons would strengthen the paper, what claims need clarification.
        \item Do \textbf{NOT} dump a bibliography; only mention the most relevant comparisons (typically $\le$ 3 items).
        \item Phrase as external suggestions, not accusations.
    \end{itemize}
\end{enumerate}

\vspace{6pt}

\noindent\textbf{Refinement checklist (do in order):}
\begin{enumerate}[left=0pt, itemsep=0pt, parsep=0pt]
    \item Fix incorrect/hallucinated statements flagged by the two audit reports.
    \item Improve Summary and Strengths with paper-grounded method + results highlights.
    \item Strengthen Weaknesses with evidence anchors and clearer critique.
    \item Add actionable Suggestions (each mapped to a weakness).
    \item Improve Questions to resolve uncertainties (especially when evidence is not found).
    \item Improve Comparative Analysis using related-work summaries with proper attribution.
    \item Ensure constructive tone and completeness across method / experiments / positioning.
\end{enumerate}

\vspace{6pt}

\noindent\textbf{Output format (JSON ONLY):}
Return a JSON object with the following keys ONLY.
\begin{itemize}[left=0pt, label={}, itemsep=0pt, parsep=0pt]
    \item Numeric fields must be numbers (not strings).
    \item \smalltt{decision} must be one of: \smalltt{"accept"}, \smalltt{"reject"}.
    \item Do not output any text outside JSON.
\end{itemize}

\begin{codebox}
{
  "summary": "...",
  "strengths": "...",
  "weaknesses": "...",
  "questions": "...",
  "soundness": 0,
  "presentation": 0,
  "contribution": 0,
  "rating": 0,
  "confidence": 0,
  "decision": "your_decision"
}
\end{codebox}

\vspace{6pt}

\noindent\textbf{Inputs:}
\begin{itemize}[leftmargin=0pt, label={}, itemsep=0pt, parsep=0pt]
 \item{Paper Text:} 
 \smalltt{<<paper\_text>>}
 \item{Draft Review:} 
 \smalltt{<<draft\_review>>}
 \item{Paper Insight Miner Output (JSON):} 
 
 \smalltt{<<insight\_miner\_json>>}
 \item{Paper Results Analyzer Output (JSON):} 
 
 \smalltt{<<results\_analyzer\_json>>}
 \item{Related-work Summaries (JSON list):} 
 
 \smalltt{<<related\_work\_json\_list>>}
\end{itemize}
\end{promptbox}

\clearpage
\section{Case Study}
% \todo{double check}

We present a qualitative case study comparing \method-generated review of paper: \textit{SEA: Sparse Linear Attention with Estimated Attention Mask}~\citep{lee2023sea}, and the one produced by \textsc{DeepReview-14B} model as baseline. Figures~\ref{fig:reviewgrounder-review} and~\ref{fig:baseline-review} present the detailed review.

\subsection{Core Contribution Identification}
Our review precisely identifies and enumerates SEA’s core technical contributions, explicitly describing the full pipeline:

\begin{reviewquotetag}{\textbf{Our review (contribution summary)}} \\
SEA first estimates a compressed $T \times K$ attention matrix using Performer-based kernel attention and a 3-layer CNN decoder, then generates a sparse mask via a novel grouped top-$\hat{k}$ selection \ldots Sparse attention is computed with a custom FlatCSR format \ldots Knowledge-distillation losses align the compressed matrix, the sparse attention, and the context features with a pretrained quadratic teacher.
\end{reviewquotetag}

This description correctly isolates the four central contributions emphasized in the paper: kernel-based estimation, grouped top-$k$ sparsification, the FlatCSR kernel, and KD-based replacement of full attention.

In contrast, the baseline review describes SEA in significantly more general terms:

\begin{reviewquotetag}{\textbf{Baseline review (generic description)}} \\
a novel approach to efficient attention mechanisms \ldots combining kernel-based linear attention with a learned sparse attention mask
\end{reviewquotetag}

Despite being broadly accurate, this baseline description omits the grouped top-$k$ mechanism as a distinct contribution and does not clearly position FlatCSR as a novel kernel design, resulting in a partial and imprecise characterization of the paper’s main innovations.

\subsection{Evidence-Based Critique}
A key strength of our review is that each critique is anchored to concrete locations in the manuscript. For example, our review writes:

\begin{reviewquotetag}{\textbf{Our review (anchored critique)}} \\
The decoder is said to be a 3-layer 2-D CNN \ldots but kernel sizes, strides, padding, and channel counts are omitted, hindering reproducibility (Section~3.1, ``CNN Decoder'').
\end{reviewquotetag}

Similarly, computational concerns are tied to specific figures:

\begin{reviewquotetag}{\textbf{Our review (anchored critique)}} \\
The latency breakdown (Fig.\ exp.figure.complexity bottom) shows percentages for dense, FlatCSR, and other ops, but absolute FLOP counts \ldots are absent.
\end{reviewquotetag}

In contrast, the baseline review’s weaknesses are largely expressed at a high level without precise anchors, e.g.,

\begin{reviewquotetag}{\textbf{Baseline review (unanchored critique)}} \\
The paper lacks a clear and detailed explanation
\end{reviewquotetag}
\begin{reviewquotetag}{\textbf{Baseline review (unanchored critique)}} \\
The paper does not provide a comprehensive analysis
\end{reviewquotetag}
\begin{reviewquotetag}{\textbf{Baseline review (unanchored critique)}} \\
The lack of a clear explanation makes it difficult to understand the method’s inner workings.
\end{reviewquotetag}

These statements are not consistently linked to specific sections, tables, or figures, making them harder for authors to rebut or act upon.

\subsection{Hallucinated External References}
More seriously, the baseline review exhibits a clear hallucination pattern by repeatedly invoking nonexistent external context, for example:

\begin{reviewquotetag}{\textbf{Baseline review (hallucinated reference)}} \\
As reviewer 1 correctly pointed out \ldots
\end{reviewquotetag}

\begin{reviewquotetag}{\textbf{Baseline review (hallucinated reference)}} \\
As reviewer 3 correctly pointed out \ldots
\end{reviewquotetag}

This disregards the fact that the document is a standalone review rather than a meta-review. Such references to other reviewers are not grounded in the paper or review setting and indicate fabricated discourse context, reducing the reliability of the critique. Our review contains no such external or fabricated references and remains self-contained throughout.

\subsection{Actionable Recommendations}
Finally, our review provides concrete, operational suggestions that can directly guide revision and follow-up work, including:

\begin{reviewquotetag}{\textbf{Our review (actionable suggestions)}} \\
Provide the full CNN decoder specification (kernel size, stride, padding, number of channels), report absolute FLOP counts for the Performer encoder, CNN decoder, and grouped top-$k$ selection, and offer guidelines for selecting the compressed width $K$ and base sparsity $k$ when adapting SEA to a new task.
\end{reviewquotetag}

These recommendations are specific, technically grounded, and directly traceable to identified weaknesses. In contrast, the baseline review’s suggestions, while numerous, remain largely generic:

\begin{reviewquotetag}{\textbf{Baseline review (generic suggestion)}} \\
Provide more detailed analysis
\end{reviewquotetag}

\begin{reviewquotetag}{\textbf{Baseline review (generic suggestion)}} \\
Expand experimental evaluation
\end{reviewquotetag}

\begin{reviewquotetag}{\textbf{Baseline review (generic suggestion)}} \\
Offer clearer explanations
\end{reviewquotetag}

Without specifying concrete experimental designs, parameter regimes, or implementation steps, these suggestions offer limited practical guidance despite their length.

% \twocolumn

\clearpage
% \todo{Note: Can we put prompt to figure env @Yi Lu}
\onecolumn
\begin{reviewboxwide}{Review from \method}
\small\ttfamily\raggedright

\textbf{Summary:}\\
The paper proposes SEA (Sparse linear Attention with Estimated Attention mask), a test-time linear-complexity attention module. SEA first estimates a compressed \(T \times K\) attention matrix using Performer-based kernel attention and a 3-layer CNN decoder, then generates a sparse mask via a novel grouped top-\(\hat{k}\) selection (per-query, per-head, per-batch, causal-per-batch) and interpolates it to a full \(T \times T\) mask. Sparse attention is computed with a custom FlatCSR format that exploits the mask structure. Knowledge-distillation losses (Eq.\ 1) align the compressed matrix, the sparse attention, and the context features with a pretrained quadratic teacher. Experiments on OPT-125M (Wikitext-2) and BERT-base (GLUE) show that SEA matches or exceeds the quadratic teacher (e.g., 0.1\% GLUE-MNLI gap, lower perplexity than the teacher on OPT-125M) while using 81\% less memory and 33\% of the latency at sequence length \(2^{13}\). Dynamic-\(k\) experiments demonstrate that increasing \(k\) after training improves accuracy and can surpass the teacher. The FlatCSR implementation yields a 6.6$\times$ speedup over a COO baseline.

\vspace{6pt}
\textbf{Strengths:}
\begin{enumerate}[left=0pt,itemsep=2pt,topsep=2pt]
  \item Introduces a two-phase linear-attention pipeline (kernel-based estimation + sparse mask) with provable \(\mathcal{O}(T)\) inference cost (Section 3.1, Fig.\ 1).
  \item Provides a concrete knowledge-distillation training scheme (Eq.\ 1) that preserves the teacher’s dynamic attention patterns.
  \item Proposes FlatCSR, a modified CSR format that leverages the grouped top-\(k\) mask, achieving up to 6.6$\times$ faster sparse operations than COO (Table 1).
  \item Empirically demonstrates state-of-the-art results on language modeling (Table baseline.opt) and GLUE (Table baseline.glue), with faster convergence (Fig.\ exp.figure.opt\_curve).
  \item Shows that the sparsity hyper-parameter \(k\) can be increased post-training to improve accuracy (Section 4.3, Fig.\ exp.figure.opt\_dynamic\_k, exp.figure.bert\_dynamic\_k).
  \item Offers interpretability via visualized estimated attention (Fig.\ exp.figure.attention) and provides the FlatCSR code.
\end{enumerate}

\vspace{6pt}
\textbf{Weaknesses:}
\begin{enumerate}[left=0pt,itemsep=2pt,topsep=2pt]
  \item \textbf{Adaptation cost not fully quantified} -- While Section 3.3 describes replacing the teacher’s attention with SEA and fine-tuning via KD, the paper does not report the additional training time, data size, or compute needed for this adaptation, which may limit plug-and-play applicability. (Section 3.3)
  \item \textbf{Limited model and task diversity} -- Experiments are confined to OPT-125M/1.3B for causal LM and BERT-base for GLUE. Broader evaluation on larger LLMs (e.g., GPT-2/3) or vision transformers would strengthen claims of generality. (Section 4)
  \item \textbf{CNN decoder architecture details missing} -- The decoder is said to be a 3-layer 2-D CNN with channel-expansion \(c_h\) and width-reduction \(c_s\), but kernel sizes, strides, padding, and channel counts are omitted, hindering reproducibility. (Section 3.1, ``CNN Decoder'')
  \item \textbf{Computational overhead not fully broken down} -- The latency breakdown (Fig.\ exp.figure.complexity bottom) shows percentages for dense, FlatCSR, and other ops, but absolute FLOP counts for the Performer encoder, CNN decoder, and top-\(k\) selection are absent, making it hard to assess the true cost of the estimator. (Fig.\ exp.figure.complexity)
  \item \textbf{Hyper-parameter selection guidance limited} -- The paper provides an ablation for grouping strategies (Table method.table.ablation\_k) and dynamic-\(k\) results, yet offers no systematic procedure for choosing \(K\) or the base \(k\) for new tasks, nor analysis of sensitivity to these choices. (Table method.table.ablation\_k, Section 4.3)
  \item \textbf{Scalability beyond \(2^{13}\) not explored} -- Memory and latency are shown up to sequence length \(2^{13}\); experiments on longer sequences (e.g., \(2^{15}\) or more) would better demonstrate linear scaling claims. (Fig.\ exp.figure.complexity)
  \item \textbf{Comparative positioning could be richer} -- Related works such as SLA (hybrid sparse-linear for diffusion models) and SPARSEK (learnable differentiable top-\(k\)) address similar efficiency goals. Direct empirical comparison or discussion of trade-offs would clarify SEA’s niche. (Related-work summaries)
\end{enumerate}

\vspace{6pt}
\textbf{Questions:}
\begin{enumerate}[left=0pt,itemsep=2pt,topsep=2pt]
  \item What is the exact training budget (GPU hours, dataset size) required to replace a pretrained transformer’s attention with SEA via the KD procedure described in Section 3.3?
  \item Can the authors provide the full CNN decoder specification (kernel size, stride, padding, number of channels per layer) to enable exact replication?
  \item How do FLOP counts for the Performer encoder, CNN decoder, and grouped top-\(k\) selection compare to a vanilla quadratic attention layer for a typical sequence length (e.g., \(T=8192\))?
  \item What guidelines do the authors recommend for selecting the compressed width \(K\) and base sparsity \(k\) when adapting SEA to a new model or task?
  \item Have the authors evaluated SEA on sequence lengths larger than \(2^{13}\) to confirm the claimed \(\mathcal{O}(T)\) scaling in practice?
  \item How would SEA perform against recent hybrid methods such as SLA or SPARSEK that also combine sparse and linear components, especially in terms of interpretability and training overhead?
\end{enumerate}

\vspace{6pt}
\textbf{Rating:} 5.5 \\
\textbf{Decision:} reject

\end{reviewboxwide}
\captionsetup{type=figure}
\captionof{figure}{Review from \method}
\label{fig:reviewgrounder-review}

\twocolumn
\clearpage

\clearpage
\onecolumn
\begin{reviewboxwide}{Review from DeepReviewer-14B}
\small\ttfamily\raggedright

\textbf{Summary:}\\
This paper introduces Sparse Linear Attention with an Estimated Attention Mask (SEA), a novel approach to attention mechanisms in Transformer models aimed at reducing computational complexity while maintaining performance. The core idea behind SEA is to approximate the full attention matrix through a combination of kernel-based linear attention and a sparse attention mask. The method begins by using a kernel-based linear attention mechanism, specifically Performer, to estimate a compressed attention matrix. This compressed matrix is then processed by a CNN-based decoder to generate a sparse attention mask, which is applied to the full attention matrix to achieve sparsity. The authors emphasize that their method allows for the estimation of attention matrices from a pre-trained teacher model with linear complexity, which is a significant advantage over traditional quadratic attention mechanisms. The training process involves knowledge distillation, where the student model learns to mimic the attention patterns of a pre-trained teacher model. The final output of the SEA layer is a combination of the compressed attention output and a weighted average pooling of the value embeddings. The authors evaluate their method on language modeling and text classification tasks, demonstrating that SEA achieves comparable or better performance than existing linear and sparse attention methods, while also maintaining interpretability of the attention matrix. The empirical results show that SEA can achieve similar performance to quadratic attention models while using significantly less memory. The authors also provide code for their FlatCSR tensor operation, which is a key component of their implementation. Overall, the paper presents a promising approach to efficient attention mechanisms, with a focus on both performance and interpretability. However, as I will discuss in the weaknesses section, there are several areas where the paper could be improved, particularly in terms of experimental validation and clarity of presentation.

\vspace{6pt}
\textbf{Strengths:}\\
I find several aspects of this paper to be particularly strong. First, the core idea of combining kernel-based linear attention with a learned sparse attention mask is a novel approach to addressing the computational challenges of full attention mechanisms. The use of a CNN-based decoder to generate the sparse attention mask is also an interesting and potentially effective technique. The authors' emphasis on maintaining the interpretability of the attention matrix is another significant strength, as this is often a trade-off in efficient attention methods. The empirical results presented in the paper are also compelling. The authors demonstrate that SEA achieves comparable or better performance than existing linear and sparse attention methods on language modeling and text classification tasks, while also using less memory. The fact that SEA can achieve similar performance to quadratic attention models while using significantly less memory is a notable achievement. Furthermore, the authors provide code for their FlatCSR tensor operation, which is a valuable contribution to the community. The ablation studies, particularly those on the different components of the loss function and the top-$k$ selection strategy, provide valuable insights into the method's behavior. The authors also show that the proposed method can be used to distill knowledge from a teacher model, which is a useful feature for practical applications. Finally, the paper is generally well-written and easy to follow, which makes it accessible to a wide audience. The inclusion of visualizations of the attention matrices also helps to understand the method's behavior. The authors have clearly put a lot of effort into developing and evaluating their method, and the results are promising.

\vspace{6pt}
\textbf{Weaknesses:}
\begin{enumerate}[left=0pt,itemsep=4pt,topsep=2pt]
  \item Despite the strengths of this paper, I have identified several weaknesses that I believe need to be addressed. First, the paper lacks a clear and detailed explanation of how the proposed method can be applied to pre-trained Transformer models. As the reviewer 1 correctly pointed out, the method requires training the attention estimation module with knowledge distillation, which means it cannot be directly applied to already pre-trained models without further training. The paper does not provide a clear strategy for adapting pre-trained models to use SEA attention, nor does it discuss the computational cost and data requirements for this adaptation process. This is a significant limitation, as it restricts the applicability of the method to scenarios where fine-tuning is feasible. The paper states that the SEA attention mechanism replaces the original attention mechanism and is trained using knowledge distillation, but it does not provide a clear explanation of how this process would work in practice, especially for large pre-trained models. This lack of clarity makes it difficult to assess the practical value of the proposed method. My confidence in this weakness is high, as the paper explicitly describes a training process involving knowledge distillation, which implies that the method is not plug-and-play for pre-trained models.

  \item Second, the paper's experimental evaluation is limited in scope. As reviewer 1 and reviewer 2 both noted, the experiments are primarily focused on BERT and OPT models, and the paper lacks experiments on other popular pre-trained models such as GPT, T5, or ViT. This is a significant limitation, as it restricts the generalizability of the findings. The paper also lacks a systematic evaluation of the method's performance on tasks that require long-range dependencies, such as long document summarization or question answering. The experiments are primarily conducted on the GLUE benchmark and Wikitext2, which do not fully capture the challenges of long sequence processing. Furthermore, the paper does not include a detailed analysis of the method's performance under different sequence lengths and batch sizes, which is crucial for understanding its scalability. The absence of these experiments makes it difficult to assess the robustness and general applicability of the proposed method. My confidence in this weakness is high, as the experimental section clearly focuses on a limited set of models and tasks, and lacks the suggested analyses.

  \item Third, the paper lacks a comprehensive analysis of the computational cost of the proposed method. While the paper provides some analysis of latency and memory usage, it does not include a detailed breakdown of the FLOPs for each component of the SEA layer, including the CNN decoder, the top-$k$ selection, and the final attention computation. As reviewer 2 correctly pointed out, the paper does not provide a detailed analysis of the computational overhead introduced by the CNN decoder and the top-$k$ selection process. This makes it difficult to assess the true computational advantage of the proposed method compared to other efficient attention mechanisms. Furthermore, the paper does not provide a comparison of the training time and memory requirements of the proposed method with other efficient attention mechanisms. This lack of analysis makes it difficult to assess the practical efficiency of the proposed method. My confidence in this weakness is high, as the paper does not provide the detailed FLOPs analysis and training cost comparison requested by the reviewer.

  \item Fourth, the paper's explanation of the CNN decoder is insufficient. As reviewer 3 correctly pointed out, the paper does not provide a clear explanation of the specific architecture of the CNN decoder, including the kernel sizes, number of channels, and padding used in each layer. The paper also does not provide a clear justification for the use of a CNN decoder, and it does not discuss the potential limitations of the CNN decoder, such as its ability to capture long-range dependencies. The paper mentions that the CNN decoder is used to transform the Performer's estimated output, but it does not explain why this transformation is necessary or how it contributes to the overall performance of the model. The lack of a clear explanation of the CNN decoder makes it difficult to understand the method's inner workings. My confidence in this weakness is high, as the paper provides a high-level description of the CNN decoder but lacks the specific details requested by the reviewer.

  \item Fifth, the paper's explanation of the grouped top-$k$ selection is also insufficient. As reviewer 3 correctly pointed out, the paper does not provide a clear explanation of the different grouping strategies and their impact on the performance of the model. The paper mentions that the top-$k$ selection is applied to the compressed attention matrix, but it does not explain how the value of $K$ is chosen or how it affects the sparsity of the attention matrix. The paper also does not provide a clear explanation of the different grouping strategies, such as per-query, per-head, per-batch, and causal-per-batch, and their impact on the performance of the model. The lack of a clear explanation of the grouped top-$k$ selection makes it difficult to understand the method's behavior. My confidence in this weakness is high, as the paper describes the grouping methods but lacks a detailed explanation of their differences and impact.

  \item Sixth, the paper's explanation of the loss functions is unclear. As reviewer 3 correctly pointed out, the paper does not provide a clear explanation of the different loss functions used for training, including the context distillation loss, attention distillation loss, and the task-specific loss. The paper also does not explain how the different loss functions are weighted and how these weights affect the performance of the model. The paper does not provide a clear explanation of the context distillation loss, and it does not explain how the attention matrices are interpolated to the same size. The lack of a clear explanation of the loss functions makes it difficult to understand the training process. My confidence in this weakness is high, as the paper describes the loss functions but lacks the detailed explanations requested by the reviewer.

  \item Finally, the paper lacks a detailed analysis of the impact of the sparsity hyperparameter $k$ on the performance of the model. As reviewer 3 correctly pointed out, the paper does not provide a detailed analysis of the trade-off between accuracy and efficiency when varying $k$. The paper also does not provide a clear explanation of how the value of $k$ should be chosen for different tasks and datasets. The paper does not provide a detailed analysis of the impact of the sparsity hyperparameter $k$ on the performance of the model, including a detailed analysis of the trade-off between accuracy and efficiency when varying $k$. The paper also does not provide a clear explanation of how the value of $k$ should be chosen for different tasks and datasets. My confidence in this weakness is high, as the paper mentions $k$ as a hyperparameter but lacks a detailed analysis of its impact. Additionally, the paper does not provide a detailed analysis of the computational cost of the proposed method, including a breakdown of the time spent on different operations, such as the CNN decoder, top-$k$ selection, and the final attention computation. The paper also does not provide a comparison of the computational cost of the proposed method with other efficient attention mechanisms. Furthermore, the paper does not provide a detailed analysis of the memory usage of the proposed method, including a breakdown of the memory used by different components of the model. The paper also does not provide a comparison of the memory usage of the proposed method with other efficient attention mechanisms. The paper also does not provide a detailed analysis of the scalability of the proposed method, including a breakdown of the performance of the method on different sequence lengths and batch sizes. The paper also does not provide a comparison of the scalability of the proposed method with other efficient attention mechanisms. Finally, the paper does not provide a detailed analysis of the robustness of the proposed method, including a breakdown of the performance of the method on different datasets and tasks. The paper also does not provide a comparison of the robustness of the proposed method with other efficient attention mechanisms. These limitations make it difficult to assess the practical applicability of the proposed method. My confidence in this weakness is high, as the paper does not provide the detailed analysis requested by the reviewer.
\end{enumerate}

\vspace{6pt}
\textbf{Suggestions:}\\
Based on the identified weaknesses, I recommend several concrete improvements. First, the authors should provide a more detailed explanation of how the proposed method can be applied to pre-trained Transformer models. This should include a clear strategy for adapting pre-trained models to use SEA attention, as well as an analysis of the computational cost and data requirements for this adaptation process. The authors should also discuss the potential impact of this adaptation process on the performance of the model. Second, the authors should expand their experimental evaluation to include a wider range of pre-trained models, such as GPT, T5, and ViT. They should also include experiments on tasks that require long-range dependencies, such as long document summarization or question answering. Furthermore, they should conduct a systematic evaluation of the method's performance under different sequence lengths and batch sizes. Third, the authors should provide a more detailed analysis of the computational cost of the proposed method. This should include a breakdown of the FLOPs for each component of the SEA layer, as well as a comparison of the training time and memory requirements of the proposed method with other efficient attention mechanisms. Fourth, the authors should provide a more detailed explanation of the CNN decoder, including the specific architecture, kernel sizes, number of channels, and padding used in each layer. They should also provide a clear justification for the use of a CNN decoder and discuss its potential limitations. Fifth, the authors should provide a more detailed explanation of the grouped top-$k$ selection, including a clear explanation of the different grouping strategies and their impact on the performance of the model. They should also explain how the value of $K$ is chosen and how it affects the sparsity of the attention matrix. Sixth, the authors should provide a more detailed explanation of the loss functions, including a clear explanation of the context distillation loss, attention distillation loss, and the task-specific loss. They should also explain how the different loss functions are weighted and how these weights affect the performance of the model. Seventh, the authors should provide a more detailed analysis of the impact of the sparsity hyperparameter $k$ on the performance of the model. This should include a detailed analysis of the trade-off between accuracy and efficiency when varying $k$, as well as a clear explanation of how the value of $k$ should be chosen for different tasks and datasets. Finally, the authors should provide a more detailed analysis of the computational cost, memory usage, scalability, and robustness of the proposed method, including a breakdown of the performance of the method on different sequence lengths, batch sizes, datasets, and tasks. They should also provide a comparison of these metrics with other efficient attention mechanisms. These improvements would significantly strengthen the paper and make it more impactful. The authors should also consider providing a more detailed analysis of the attention patterns learned by the proposed method, and how they compare to the attention patterns learned by the teacher model. This would help to better understand the behavior of the proposed method and its potential limitations. The authors should also consider providing a more detailed analysis of the sensitivity of the proposed method to different hyperparameter settings. This would help to better understand the robustness of the proposed method and its potential for practical applications.

\vspace{6pt}
\textbf{Questions:}\\
Based on my analysis, I have several questions that I believe would be beneficial for the authors to address. First, given that the proposed method requires training the attention estimation module with knowledge distillation, what is the computational cost and data requirement for adapting a pre-trained Transformer model to use SEA attention? What is the expected performance degradation, if any, when applying the method to a pre-trained model compared to training a model from scratch with SEA attention? Second, what is the specific architecture of the CNN decoder, including the kernel sizes, number of channels, and padding used in each layer? What is the rationale behind the choice of this specific architecture, and what are the potential limitations of the CNN decoder? Third, what is the impact of the different grouping strategies in the grouped top-$k$ selection on the performance of the model? How should the value of $K$ be chosen for different tasks and datasets, and what is the trade-off between accuracy and efficiency when varying $K$? Fourth, what is the specific purpose of the context distillation loss, and how does it contribute to the overall performance of the model? How are the different loss functions weighted, and how do these weights affect the performance of the model? Fifth, what is the computational cost of the proposed method, including a breakdown of the time spent on different operations, such as the CNN decoder, top-$k$ selection, and the final attention computation? How does the computational cost of the proposed method compare to other efficient attention mechanisms? Sixth, what is the memory usage of the proposed method, including a breakdown of the memory used by different components of the model? How does the memory usage of the proposed method compare to other efficient attention mechanisms? Seventh, how does the proposed method scale with increasing sequence lengths and batch sizes? What are the limitations of the proposed method in terms of scalability? Finally, how robust is the proposed method to different datasets and tasks? What are the potential limitations of the proposed method in terms of robustness? Addressing these questions would provide a more complete understanding of the proposed method and its potential for practical applications.

\vspace{6pt}
\textbf{Rating:} 5.5\\
\textbf{Decision:} Reject

\end{reviewboxwide}
\captionsetup{type=figure}
\captionof{figure}{Review from DeepReviewer-14B}
\label{fig:baseline-review}

\twocolumn
\clearpage

\end{document}